\documentclass[11pt]{article}

\usepackage[preprint]{acl}

\usepackage{enumitem}
\usepackage{times}
\usepackage{latexsym}
\usepackage{amsmath}
\usepackage{graphicx}   
\usepackage{subcaption} 
\usepackage[most]{tcolorbox}
\tcbuselibrary{skins, breakable}
\usepackage{fontawesome}
\usepackage{hyperref}
\usepackage{url}
\usepackage{amssymb} 
\usepackage{booktabs}
\usepackage{algorithm}
\usepackage{algpseudocode}
\usepackage{amsmath}
\usepackage{amssymb}
\usepackage{enumitem}

\usepackage{multirow}
\usepackage[table]{xcolor}
\usepackage{threeparttable}
\usepackage{makecell}

\usepackage[T1]{fontenc}
\usepackage{booktabs}
\usepackage{tabularx}

\definecolor{Highlight}{rgb}{0.92,0.94,1}
\definecolor{cream}{RGB}{255, 251, 234}
\definecolor{goldenstar}{RGB}{255, 204, 0}
\usepackage{tikz}
\usepackage{xcolor}

\usepackage[T1]{fontenc}

\usepackage[utf8]{inputenc}
\usepackage{microtype}

\usepackage{inconsolata}
\usepackage{graphicx}

\title{Towards Fair and Comprehensive Evaluation of Routers \\in Collaborative LLM Systems}

\usepackage{tcolorbox}
\tcbuselibrary{skins, breakable}  

\definecolor{cream}{RGB}{255, 253, 245}
\definecolor{goldenstar}{RGB}{255, 193, 7}

\tcbset{
  takeawaystyle/.style={
    colback=Highlight,
    colframe=black!30,
    boxrule=0.4pt,
    arc=3pt,
    left=6pt,
    right=6pt,
    top=6pt,
    bottom=6pt,
    fonttitle=\bfseries,
    coltitle=black,
    enhanced,
    drop shadow
  },
  promptstyle/.style={  
    colback=blue!5!white,
    colframe=blue!75!black,
    boxrule=0.5pt,
    arc=2pt,
    left=8pt,
    right=8pt,
    top=8pt,
    bottom=8pt,
    fonttitle=\bfseries\sffamily,
    coltitle=white,
    enhanced,
    attach boxed title to top left={yshift=-2mm, xshift=5mm},
    boxed title style={
      colback=blue!75!black,
      arc=2pt,
    },
  }
}

\author{%
    Wanxing Wu$^{1,2*}$, \
    He Zhu$^{3*}$, \
    Yixia Li$^{1}$\thanks{\ \ Equal Contributions.}, \
    Lei Yang$^{4}$,  \
    Jiehui Zhao$^{4}$ \ \\
    \textbf{Hongru Wang}$^{5}$,  \
    \textbf{Jian Yang}$^{6}$,  \
    \textbf{Benyou Wang}$^{7}$,  \
    \textbf{Bingyi Jing}$^{7}$,  \
    \textbf{Guanhua Chen}$^{1}$\thanks{\ \ Corresponding author.} \\
    $^1$Southern University of Science and Technology,
    $^2$Institut Polytechnique de Paris\\ 
    $^3$Peking University,
    $^4$Deepexi Technology Co. Ltd.,
    $^5$University of Edinburgh\\
    $^6$Beihang University, 
    $^7$Chinese University of Hong Kong (Shenzhen) 
}

\usepackage{xcolor}

\begin{document}
\maketitle
\begin{abstract}
  Large language models (LLMs) have achieved success, but cost and 
privacy constraints necessitate deploying smaller models locally while 
offloading complex queries to cloud-based models. Existing router evaluations are unsystematic, overlooking scenario-specific requirements and out-of-distribution robustness. We propose \textbf{RouterXBench}, a principled 
evaluation framework with three dimensions: router ability, scenario 
alignment, and cross-domain robustness. Unlike prior work that relies on output probabilities or external embeddings, we utilize internal hidden states that capture model uncertainty before answer generation. We introduce \textbf{ProbeDirichlet}, a lightweight router that aggregates cross-layer hidden states via learnable 
Dirichlet distributions with probabilistic training. 
Trained on multi-domain data, it generalizes robustly 
across in-domain and out-of-distribution scenarios. Our results show ProbeDirichlet achieves 16.68\% and 18.86\% relative improvements over the best baselines in router ability and high-accuracy scenarios, with consistent performance across model families, model scales, heterogeneous tasks, and agentic workflows.
\end{abstract}

\section{Introduction}

Large Language Models (LLMs) achieve remarkable performance across diverse 
tasks such as language understanding, creative writing, and code generation\citep{zhao2023surveyllm,matarazzo2025surveyllm}, but balancing cost and accuracy under varying deployment constraints remains 
a key challenge. Routers address this by dynamically directing queries 
to different models: routing complex queries to powerful cloud models while 
processing simpler ones on local edge devices\citep{dingHybridLLMCostEfficient2024, zhang2025leveraginguncertaintyestimationefficient, barrak2025cargoframeworkconfidenceawarerouting}. 
This reduces computational cost, but may sacrifice 
some accuracy\citep{kassem2025robustrouterllmsanalysisfragility, shafran2025reroutingllmrouters, lin2025lifecycleroutingvulnerabilitiesllm}. 

However, this trade-off is not equally acceptable across domains. 
Different domains have different tolerances: safety-critical applications like 
healthcare require high reliability \citep{buschCurrentApplicationsChallenges2025c}, 
while customer support may tolerate accuracy drops for cost savings 
\citep{yu2025efficientroutinginferencerequests}.
Beyond domain-specific requirements, routers must also handle queries from 
unfamiliar distributions (out-of-distribution, OOD). Given these diverse 
requirements, a single metric cannot capture router quality. Fair evaluation 
requires assessing both deployment scenarios and cross-domain robustness.

Existing benchmarks fail to achieve this comprehensive assessment. Current evaluations rely on single metrics such as static thresholds \citep{chen2024frugalgpt,ding2024hybridllmcostefficientqualityaware,stripelis2024tensoroperaroutermultimodelrouter,aggarwalAutoMixAutomaticallyMixing2024} or curve-based aggregate scores\citep{rez2024optimising,hu2024routerbench,ong2025routellm}, which cannot capture the multifaceted trade-offs required across diverse application scenarios (Subsection~\ref{Current Metrics}). Beyond metric limitations, many studies evaluate routing performance solely on in-distribution data without systematic out-of-distribution (OOD) assessment. However, real-world deployments face diverse, shifting query distributions, requiring comprehensive evaluation of both scenario-specific performance and cross-domain robustness.

Motivated by these gaps, we propose \textbf{RouterXBench} a systematic evaluation framework 
spanning three key dimensions: 
\textbf{(i) Router Ability}, measured by AUROC to capture a router's 
fundamental discrimination capability independent of deployment thresholds; 
\textbf{(ii) Scenario Alignment}, quantified by metrics tailored to 
low-cost, balanced, and high-accuracy deployment regimes (detailed in 
Section~\ref{sec:metrics}); 
and \textbf{(iii) Cross-Domain Robustness}, assessed across diverse 
in-distribution (ID) and out-of-distribution (OOD) tasks. 
By disentangling intrinsic routing ability from scenario-specific 
requirements, our framework enables more principled router comparison and guides our 
exploration of effective routing design.

We then focus on the core challenge: \textbf{How to construct routing that 
is both effective and generalizable?} We explore router design and training 
data composition, validated on our evaluation framework and agentic applications. 
Internal hidden states directly capture model uncertainty before answer generation, 
proving more reliable than output probabilities that suffer from softmax 
overconfidence~\citep{guo2017calibration}. To robustly aggregate cross-layer 
representations, we model layer importance using a Dirichlet distribution with 
learned concentration parameters. This enables stochastic training with 
deterministic inference, acting as layer dropout to prevent overfitting specific 
layers. We show that diverse data mixtures improve cross-domain generalization 
while preserving in-distribution performance. Our approach achieves 16.68\% and 
18.86\% relative improvements in router ability and HCR over state-of-the-art 
baselines, with strong generalization across model families, 
model scales, diverse scenarios, and agentic workflows.\footnote{Our code is publicly available at \url{https://github.com/zhuchichi56/RouterXBench}.}

\section{Related Work}

\paragraph{LLM Routing.}  
Prior work explores several technical directions. Training-free approaches avoid labeled supervision by estimating model skill from relative performance \citep{zhao2024eagleefficienttrainingfreerouter} or leveraging weak agreement signals \citep{guha2024smoothie,aggarwalAutoMixAutomaticallyMixing2024}. Learning-based routing methods train models to predict which model should 
handle each query, including preference-based routers \citep{ong2025routellm}, contrastive query–model embedding alignment \citep{chen2024routerdc}, and instruction-level capability encoding \citep{zhang2025capabilityinstructiontuningnew}. Adaptive routing formulates routing as sequential decision making, such as bandit-based selection \citep{li2025llm} or token-level deferral from small to large models \citep{she2025tokenlevelroutinginference}. Quality- and compute-aware designs integrate routing with explicit test-time budget control, such as Hybrid LLM \citep{ding2024hybridllmcostefficientqualityaware} and BEST-Route \citep{ding2025bestroute}. Beyond specific router designs, recent benchmarking efforts such as RouterEval \citep{hu2025routereval} provide comprehensive frameworks to evaluate routing performance and explore the scaling effects of integrating multiple models of varying capacities.

\paragraph{LLM Collaboration.}  
Collaboration strategies complement routing by coordinating multiple models or agents. Representative directions include speculative decoding, which accelerates inference using a draft–verifier pair \citep{chen2023acceleratinglargelanguagemodel,cai2024medusasimplellminference,li2024eagle2fasterinferencelanguage}, and model cascades, which escalate queries through models of increasing capacity with calibrated deferral rules \citep{chen2024frugalgpt,gupta2024language}. More recent work explores multi-agent systems with specialized roles and coordination protocols \citep{wu2024autogen,li2023camel,wang2025anymaccascadingflexiblemultiagent}.

\paragraph{LLM Uncertainty Estimation.}  
Uncertainty estimation provides key signals for routing. Existing methods include information-based scores such as perplexity or entropy \citep{fomicheva-etal-2020-unsupervised,duan2024shifting,fadeeva-etal-2024-fact}, consistency-based signals from agreement across generations \citep{kuhn2023semantic,lin2024generating,qiu2024semantic}, and introspective probes using hidden states or attention patterns \citep{chen2024inside,sriramanan2024llmcheck,lin-etal-2024-contextualized}. These methods can be integrated into routers to improve decision reliability, though many were originally developed outside the routing context.

\section{Evaluation Framework}

\subsection{Problem Setup}
\label{sec:metrics_definition}

We consider routing between two models in an \textit{edge–cloud collaboration} setting:  
a small model $\mathcal{M}_{\text{small}}$ deployed locally on edge devices for low latency and privacy,  
and a large model $\mathcal{M}_{\text{large}}$ deployed in the cloud for higher accuracy at greater cost.
Given a query $q \in \mathcal{Q}$, the router decides which model to invoke.  
Let $\delta_{\text{small}}(q), \delta_{\text{large}}(q) \in [0,1]$ denote the performance of the two models on $q$.  
The router computes a score $s(q)\!\in\!\mathbb{R}$, and the decision is made by thresholding:  
\begin{equation}
r(q;\theta) = \mathbf{1}\{s(q)\geq \theta\},
\end{equation}
where $r(q;\theta)=1$ routes to the large model and $r(q;\theta)=0$ uses the small model.  
The resulting system performance under threshold $\theta$ is
\begin{equation}
\delta(q;\theta) = (1-r(q;\theta))\,\delta_{\text{small}}(q) 
                 + r(q;\theta)\,\delta_{\text{large}}(q).
\end{equation}

For a given threshold, the large-model call rate is  
\begin{equation}
d(\theta)\;=\;\frac{1}{|\mathcal{Q}|}\sum_{q\in\mathcal{Q}} r(q;\theta)\;\in[0,1],
\end{equation}
and the corresponding overall performance is  
\begin{equation}
\mathrm{Perf}(\theta)\;=\;\frac{1}{|\mathcal{Q}|}\sum_{q\in\mathcal{Q}} \delta(q;\theta).
\end{equation}

Varying the threshold $\theta$ traces out the \emph{cost–performance curve}:
\begin{equation}
\Phi:\; d(\theta)\;\mapsto\;\mathrm{Perf}(\theta).
\end{equation}
Since $d(\theta)$ is monotonic, we re-parameterize this curve as a continuous function $\Phi(x)$ of the call rate $x \in [0,1]$ via linear interpolation, which serves as the basis for our integral metrics.

\begin{figure}[t]
    \centering
    \includegraphics[width=0.49\linewidth]{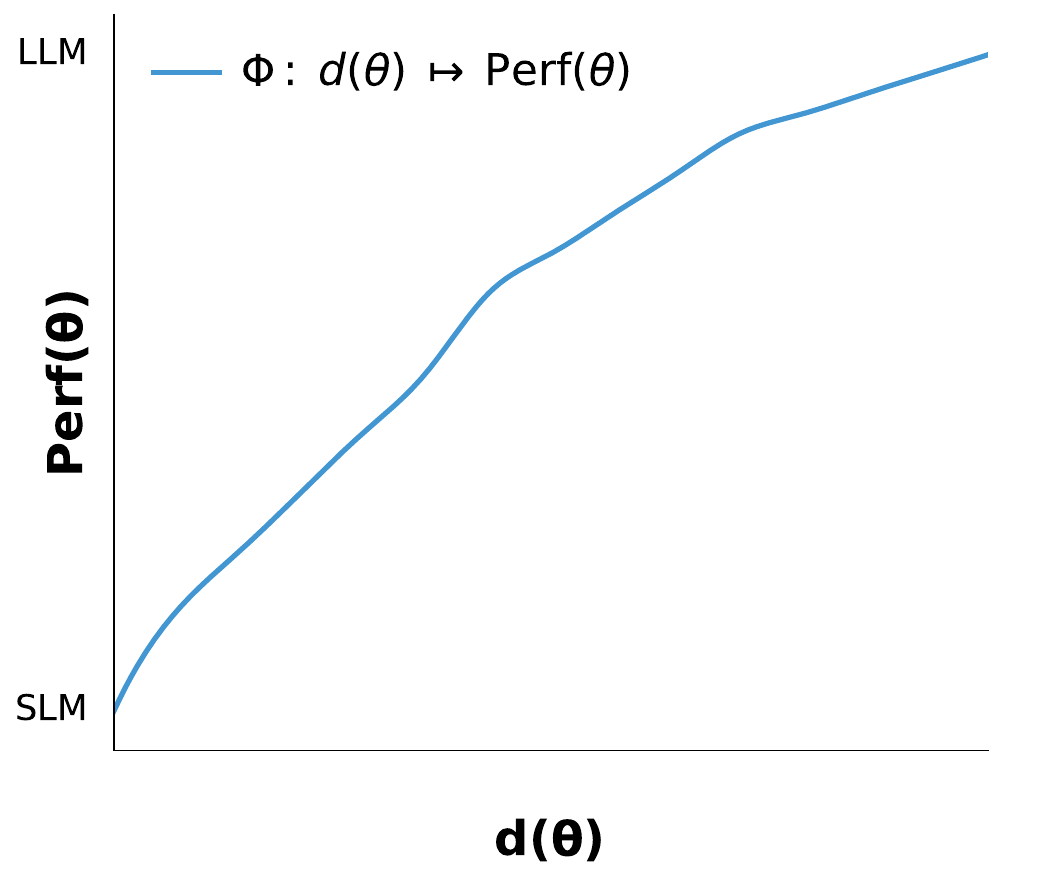}
    \hfill
    \includegraphics[width=0.49\linewidth]{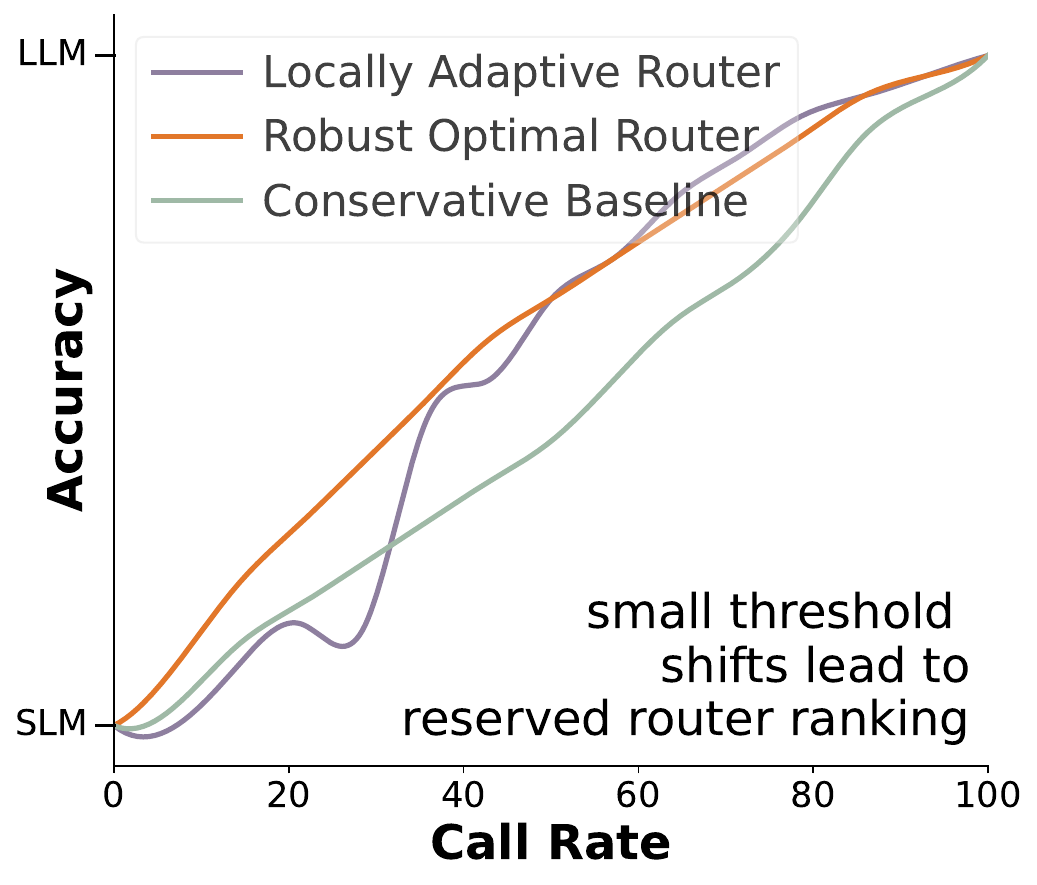}
    \caption{%
        \textbf{Left:} Cost--performance mapping where $d(\theta)$ represents 
        the call rate at threshold $\theta$ and $\text{Perf}(\theta)$ denotes 
        overall performance. By varying $\theta$, this can be re-parameterized 
        as call rate vs.\ performance (see \S\ref{sec:metrics_definition}). 
        \textbf{Right:} An illustrative limitation of existing metrics.
    }
    \label{fig:routing_metrics}
\end{figure}
\subsection{Limitations of Current Metrics}
\label{Current Metrics}
As shown in Figure~\ref{fig:routing_metrics}(left), the cost–performance curve introduced above provides a unified view of router behavior. Existing metrics can be seen as different ways of extracting information from this curve, which broadly fall into two categories. 

\paragraph{Static Metrics.}
These methods evaluate routers at fixed thresholds or compress performance into few indicators. A common approach is the cost–accuracy trade-off: FrugalGPT \citep{chen2024frugalgpt} fixes accuracy and reports cost savings, while HybridLLM \citep{ding2024hybridllmcostefficientqualityaware} fixes cost and measures accuracy drop. Others use single or composite indicators. TO-Router \citep{stripelis2024tensoroperaroutermultimodelrouter} reports total inference cost, throughput, semantic similarity, and negative log-likelihood. AutoMix \citep{aggarwalAutoMixAutomaticallyMixing2024} uses Incremental Benefit per Cost, normalizing accuracy improvement by cost into a single score.  

\textit{Limitation.}
\label{sec:metrics_limits}
While static metrics are simple and interpretable, they provide only a fragmented view of router behavior. As illustrated in Figure~\ref{fig:routing_metrics} (right), router rankings can be highly sensitive to threshold choice: within the call-rate range 20\% to 40\% , even minor shifts can lead to opposite conclusions about the Locally Adaptive Router, indicating that static evaluations may capture incidental fluctuations rather than a router’s consistent behavior.

\paragraph{Curve-based Metrics.}
These methods integrate performance over the entire cost–performance curve to avoid thresholds. Examples include the AUC (area under the accuracy–cost curve) \citep{rez2024optimising}, Average Improvement in Quality \citep{hu2024routerbench}, and Average Performance Gap Recovered \citep{ong2025routellm}. By summarizing global trends, these metrics provide threshold-independent evaluations of the trade-off surface. 

\textit{Limitation.} Aggregation, however, is scenario-blind.
The Figure~\ref{fig:routing_metrics}(right) also shows the limitation. Locally Adaptive Router performs poorly in low call-rate regions, but AUC scores conceal this difference and limit interpretability.

More fundamentally, cost–accuracy metrics entangle two factors: \emph{router ability}, referring to the correctness of judgments relative to the small model’s capacity, and \emph{scenario alignment}, concerning the leverage of the large model’s performance. Since end-to-end accuracy at a given cost reflects both, high scores may stem from the large model’s strength rather than the router’s skill, preventing faithful assessment of intrinsic routing capability.

\begin{figure*}[t]
    \centering
    \includegraphics[width=\linewidth]{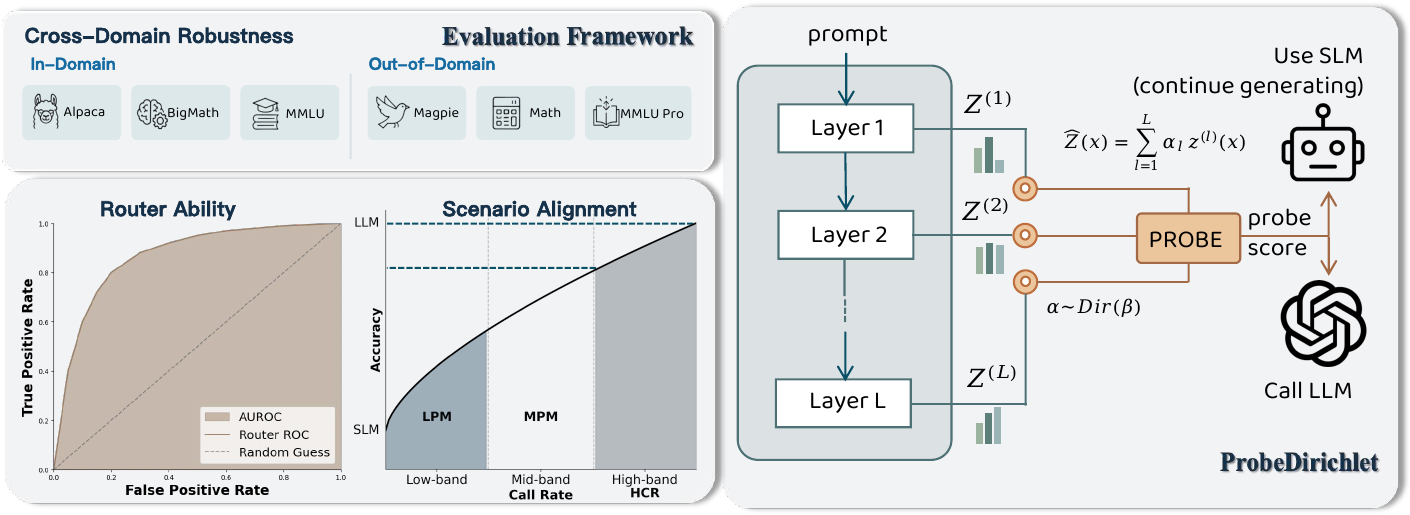}
    \caption{
        Overview of the ProbeDirichlet router and RouterXBench evaluation framework.
        Router ability is quantified using AUROC, measuring the router's accuracy in predicting whether the SLM can answer correctly. Scenario alignment is evaluated across three call-rate regimes: low band (Low-band Performance Mean, LPM), mid band (Mid-band Performance Mean, MPM), and high band (High-band Call-Rate, HCR).    
    }
    \label{fig:framework}
\end{figure*}

\subsection{Triple-Perspective Framework}
\label{sec:metrics}
To address this conflation, we propose a triple-perspective framework, \textbf{RouterXBench} (Figure~\ref{fig:framework}), that independently evaluates three distinct dimensions of routing performance. AUROC captures intrinsic discriminative ability without considering deployment costs. LPM, HCR, and MPM assess scenario alignment by quantifying how well routing matches specific cost-quality constraints. Cross-domain robustness examines performance stability across diverse task 
distributions to ensure reliable generalization.

\paragraph{1. Router Ability.} 
Since the router's primary role is to decide which model to invoke, end-to-end system accuracy may blur its individual contribution. 
To isolate the router's discriminative power from the large model's 
capabilities, we define ground truth labels based on the small 
model's performance.
Varying the decision threshold traces an ROC curve, and the area under this curve (AUROC) provides a threshold-independent measure of discriminative ability. 
Unlike cost-accuracy metrics, it focuses solely on the router's decision quality, and by aggregating over all thresholds, it avoids sensitivity to local fluctuations or opportunistic peaks.

\paragraph{2. Scenario Alignment.}  
Routers with similar intrinsic ability can behave differently under deployment constraints. 
To reflect such differences, we partition the cost–performance curve into three regions:  
(i) low call-rate for budget-sensitive use,  
(ii) high accuracy for safety-critical domains, and  
(iii) a middle band for balanced deployment.  
For each region, we define a normalized mean metric: LPM, HCR, and MPM. As illustrated in Figure \ref{fig:framework}.

\textbf{Low-band Performance Mean (LPM).} 
For strict budget scenarios, let $d_1 \in (0,1]$ denote the maximum allowable call rate.
The average performance in this region is defined as:
\begin{equation}
\mathrm{LPM} = \frac{1}{d_1}\int_{0}^{d_1}\Phi(x)\,dx.
\end{equation}

\textbf{High-band Call Rate (HCR).}
For accuracy-critical applications, we target a specific Relative Performance (RP) range.
Given an RP interval $[\rho_1, \rho_2]$, we map these to absolute performance thresholds $[\tau_1, \tau_2]$ via:
\begin{equation}
\tau_i = \mathrm{Perf}_S + \rho_i (\mathrm{Perf}_L - \mathrm{Perf}_S), \quad i \in \{1, 2\}.
\end{equation}
We then identify the \emph{feasible call-rate set} $\mathcal{D}$ where the router's performance curve $\Phi(x)$ falls within this absolute band:
\begin{equation}
\mathcal{D} = \{x \in [0,1] : \tau_1 \le \Phi(x) \le \tau_2\}.
\end{equation}
The HCR metric computes the complement of the average call rate within this feasible set:
\begin{equation}
\mathrm{HCR} = 1 - \frac{1}{|\mathcal{D}|}\int_{x \in \mathcal{D}} x \, dx.
\end{equation}
A higher HCR indicates the router maintains high accuracy while relying more on the small model.

\textbf{Mid-band Performance Mean (MPM).} 
This metric evaluates the trade-off efficiency in the transition region between the strict budget constraint ($d_1$) and the accuracy-critical zone. 
Let $d_2$ be the minimum call rate required to satisfy the high-accuracy threshold $\tau_1$:
\begin{equation}
d_2 = \min \{x \in [0,1] : \Phi(x) \ge \tau_1\}.
\end{equation}
The mid-band interval is defined as $(d_1, d_2]$. 
Provided that a valid transition region exists, the mean performance is:
\begin{equation}
\mathrm{MPM} = \frac{1}{d_2 - d_1} \int_{d_1}^{d_2} \Phi(x) \, dx.
\end{equation}

\paragraph{3. Cross-Domain Robustness}
We assess cross-domain robustness by evaluating Router Ability across multiple in-distribution (ID) and out-of-distribution (OOD) pairs. This presentation highlights how routers generalize to diverse domains, with benchmarks fully described in Subsection~\ref{sec:exp_benchmarks}.

\section{Methodology}
Guided by this framework, we explore three key aspects: routing on internal hidden states, cross-layer aggregation, and diverse training data.

\paragraph{Motivation.} 
A key challenge in router design is achieving robust performance across 
both in-distribution and out-of-distribution scenarios. Recent studies reveal 
that existing routing systems suffer from notable performance degradation 
under distribution shifts~\citep{ong2025routellm,hu2025routereval}. These 
approaches primarily rely on output-based features 
\citep{aggarwalAutoMixAutomaticallyMixing2024,zhang2025leveraginguncertaintyestimationefficient} 
or external embedding models\citep{feng2025fusionfactory} to assess query 
difficulty. 
\begin{table*}[!t]
    \centering
    \caption{Router ability (AUROC) comparison of routing strategies across multiple benchmarks.}
    \small
    \resizebox{\textwidth}{!}{
    \begin{tabular}{l|cccc|ccccccc}
    \toprule
    
    \multirow{2}{*}{\textbf{Method}}
    & \multicolumn{4}{c|}{\textbf{In-Domain}}
    & \multicolumn{7}{c}{\textbf{Out-of-Domain}} \\
    \cmidrule(lr){2-5} \cmidrule(lr){6-12}
    & Alpaca & Big Math & MMLU & \textbf{AVG}
    & Magpie & MATH & STEM & Human. & Social Sci. & Others & \textbf{AVG} \\
    \midrule
    SelfAsk & 49.03 & 47.20 & 53.75 & 49.99 & 37.09 & 49.29 & 53.74 & 55.86 & 56.06 & 50.91 & 50.49 \\
    SemanticEntropy & 62.02 & 55.81 & 53.93 & 57.25 & 58.82 & 55.25 & 56.27 & 51.72 & 52.90 & 53.95 & 54.82 \\
    ConfidenceMargin & 53.38 & 56.18 & 46.56 & 52.04 & 43.08 & 50.05 & 54.42 & 46.97 & 54.37 & 49.52 & 49.73 \\
    Entropy & 46.24 & 51.41 & 49.26 & 48.97 & 52.62 & 55.30 & 49.70 & 52.36 & 48.54 & 49.23 & 51.29 \\
    MaxLogits & 57.96 & 47.39 & 43.82 & 49.72 & 60.86 & 47.00 & 50.03 & 50.53 & 41.14 & 46.43 & 49.33 \\
    EmbeddingMLP & 67.31 & 56.18 & 54.89 & 59.46 & 68.97 & 56.97 & 52.97 & 53.77 & 48.16 & 50.45 & 55.22 \\
    \cellcolor{Highlight}ProbeDirichlet 
    & \cellcolor{Highlight}\textbf{72.02} & \cellcolor{Highlight}\textbf{66.18} & \cellcolor{Highlight}\textbf{67.88} & \cellcolor{Highlight}\textbf{68.70}
    & \cellcolor{Highlight}\textbf{74.08} & \cellcolor{Highlight}\textbf{73.90} & \cellcolor{Highlight}\textbf{65.32} & \cellcolor{Highlight}\textbf{57.84}
    & \cellcolor{Highlight}\textbf{58.82} & \cellcolor{Highlight}\textbf{62.77} & \cellcolor{Highlight}\textbf{65.46} \\
    \bottomrule
    \end{tabular}
    }
    \label{tab:perf_char}
    \end{table*}

We argue for a different approach: routing on \textbf{internal 
hidden states} from the model itself. Unlike output signals or external embeddings, internal representations reflect uncertainty and intermediate reasoning before final answers. This enables robust routing with lightweight linear classifiers and superior cross-domain generalization.

\paragraph{Cross-layer hidden states provide fine-grained discriminative information.}
External encoders lack model-internal access, while final output probabilities 
suffer from overconfidence due to softmax normalization \citep{guo2017calibration}. 
We instead route on cross-layer hidden states.

Different layers capture complementary information: early layers encode surface 
patterns, while deeper layers represent semantic understanding 
\citep{sun2025transformer}. Relying solely on the final layer discards 
intermediate uncertainty. Moreover, internal representations encode task 
difficulty before answer generation \citep{dong2025emergent}. We therefore 
extract and aggregate hidden states directly after the query prefix, combining 
cross-layer richness with computational efficiency.

\paragraph{Dirichlet Aggregation: Probabilistic Training, Deterministic Inference.}
As shown in Figure~\ref{fig:framework}, we first extract sentence-level representations by mean pooling over token-wise 
hidden states at each layer $l$:
\begin{equation}
z^{(l)}(x) = \frac{1}{T} \sum_{t=1}^{T} h_{t}^{(l)}.
\label{eq:mean_pooling}
\end{equation}
The final representation aggregates across layers via a weighted combination:
\begin{equation}
\hat z(x) = \sum_{l=1}^{L} \alpha_l z^{(l)}(x).
\label{eq:aggregation}
\end{equation}
\noindent\textbf{Why Dirichlet?} Fixed layer weights (e.g., uniform averaging) cannot adapt to varying query complexity. Simple learned scalars $\alpha_l$ risk overfitting specific layers, especially under distribution shift. We instead introduce a \emph{probabilistic aggregation mechanism} that samples layer weights from a learned distribution during training while maintaining efficient deterministic inference.

Concretely, we learn global concentration parameters $\beta = [\beta_1, \ldots, \beta_L]$ that are shared across all inputs. During training, layer weights are sampled from a Dirichlet distribution:
\begin{equation}
\alpha \sim \mathrm{Dir}(\beta),
\label{eq:dirichlet_sampling}
\end{equation}
where larger $\beta_l$ indicates higher confidence in layer $l$'s relevance. This stochastic sampling acts as a form of \emph{layer dropout}, preventing the model from over-relying on a narrow subset of layers and encouraging robust aggregation across the entire hidden hierarchy.

During inference, we use the deterministic expected value:
\begin{equation}
\bar \alpha_l = \mathbb{E}[\alpha_l] = \frac{\beta_l}{\sum_{j=1}^{L} \beta_j}.
\label{eq:expectation}
\end{equation}
This yields a fixed set of layer weights independent of the input, eliminating both sampling overhead and network computation at test time. Intuitively, $\beta_l$ encodes the learned importance of each layer, with the Dirichlet sampling during training providing regularization that prevents over-reliance on any specific layer combination. The \textbf{Mean Pooling} variant emerges as a special case with uniform priors ($\beta_l \equiv c$ for all $l$).

\paragraph{Diverse Training Data for Cross-Domain Robustness.}
Beyond architecture design, training data composition critically impacts 
cross-domain robustness. Single-domain training encourages the router to 
exploit domain-specific patterns rather than generalizable difficulty signals, 
limiting transfer to unseen domains.

We therefore adopt a multi-domain training strategy, training across 
multiple domains simultaneously. This forces the 
router to learn cross-domain difficulty signals—such as reasoning depth or 
context length—rather than domain-specific artifacts, enabling robust transfer 
to unseen distributions.

\renewcommand{\arraystretch}{1.2}
\begin{table*}[t]
\centering
\caption{Scenario alignment ability of routing strategies across multiple benchmarks.}
\small
\resizebox{\textwidth}{!}{
\begin{tabular}{lccccccccccc}
\toprule
\multirow{2}{*}{\textbf{Method}}
& \multicolumn{4}{c}{\textbf{In-Domain}}
& \multicolumn{7}{c}{\textbf{Out-of-Domain}} \\
\cmidrule(lr){2-5} \cmidrule(lr){6-12}
& Alpaca & Big Math & MMLU & AVG
& Magpie & MATH & STEM &  Human. & Social Sci.& Others & AVG \\
\midrule
\multicolumn{12}{c}{\textit{LPM (Low Performance Mean)}} \\

SelfAsk & \textbf{76.52} & 74.10 & 77.52 & 76.05 & 63.35 & 61.46 & 57.01 & 50.58 & \textbf{59.20} & 59.99 & 58.60 \\
SemanticEntropy & 76.49 & 74.82 & 75.90 & 75.74 & 63.08 & 61.63 & 57.15 & 49.42 & 57.40 & 59.85 & 58.09 \\
ConfidenceMargin & 76.37 & 76.18 & 75.70 & 76.08 & 62.60 & 62.72 & 56.64 & 49.50 & 58.81 & 58.60 & 58.15 \\
Entropy & 76.16 & 75.32 & 75.29 & 75.59 & 63.08 & 63.81 & 55.58 & 50.77 & 57.10 & 59.18 & 58.25 \\
MaxLogits & 75.99 & 74.88 & 75.03 & 75.30 & 63.13 & 61.16 & 56.07 & 51.19 & 55.24 & 58.48 & 57.55 \\
EmbeddingMLP & 76.16 & 75.25 & 75.90 & 75.77 & 62.66 & 63.95 & 56.78 & 50.26 & 56.38 & 59.01 & 58.17 \\
\cellcolor{Highlight}ProbeDirichlet 
& \cellcolor{Highlight}76.50 & \cellcolor{Highlight}\textbf{78.82} & \cellcolor{Highlight}\textbf{78.51} & \cellcolor{Highlight}\textbf{77.95}
& \cellcolor{Highlight}\textbf{63.53} & \cellcolor{Highlight}\textbf{69.24} & \cellcolor{Highlight}\textbf{59.20}
& \cellcolor{Highlight}\textbf{51.74} & \cellcolor{Highlight}59.12 & \cellcolor{Highlight}\textbf{62.42} & \cellcolor{Highlight}\textbf{60.88} \\
\midrule
\multicolumn{12}{c}{\textit{MPM (Middle Performance Mean)}} \\
\midrule
SelfAsk & \textbf{82.04} & 81.40 & 83.92 & 82.45 & \textbf{71.91} & 75.34 & 69.47 & 62.94 & \textbf{69.66} & 70.41 & 69.95 \\
SemanticEntropy & 81.88 & 82.07 & 82.44 & 82.13 & 70.84 & 76.24 & 69.64 & 61.64 & 67.71 & 70.10 & 69.36 \\
ConfidenceMargin & 81.84 & 83.01 & 82.34 & 82.39 & 71.34 & 77.26 & 69.47 & 61.87 & 68.36 & 69.20 & 69.58 \\
Entropy & 81.74 & 82.04 & 82.25 & 82.01 & 71.65 & 77.84 & 68.79 & 63.01 & 67.69 & 69.81 & 69.80 \\
MaxLogits & 81.60 & 82.12 & 81.61 & 81.78 & 71.63 & 76.63 & 68.43 & 62.15 & 66.13 & 69.11 & 69.01 \\
EmbeddingMLP & 81.93 & 82.51 & 82.61 & 82.35 & 71.63 & 78.18 & 69.29 & 62.53 & 67.15 & 69.60 & 69.73 \\

\cellcolor{Highlight}ProbeDirichlet 
& \cellcolor{Highlight}81.96 & \cellcolor{Highlight}\textbf{84.67 }& \cellcolor{Highlight}\textbf{84.31} & \cellcolor{Highlight}\textbf{83.65}
& \cellcolor{Highlight}71.77 & \cellcolor{Highlight}\textbf{81.45} & \cellcolor{Highlight}\textbf{71.06}
& \cellcolor{Highlight}\textbf{64.51 }& \cellcolor{Highlight}69.16 & \cellcolor{Highlight}\textbf{71.73} & \cellcolor{Highlight}\textbf{71.61} \\
\midrule
\multicolumn{12}{c}{\textit{HCR (High-band Call Rate)}} \\
\midrule
SelfAsk & 10.50 & 6.00 & 12.50 & 9.67 & 13.50 & 11.50 & 13.64 & 10.75 & 11.00 & 11.83 & 12.04 \\
SemanticEntropy & \textbf{14.00} & 16.00 & 15.50 & 15.17 & \textbf{14.50} & 13.00 & \textbf{16.00} & 10.75 & 13.33 & 12.50 & 13.35 \\
ConfidenceMargin & 9.50 & 14.00 & 10.00 & 11.17 & 9.50 & 10.00 & 12.50 & \textbf{12.25} & 11.68 & 8.17 & 10.68 \\
Entropy & 11.50 & 8.50 & 9.00 & 9.67 & 11.00 & 12.50 & 8.83 & 9.23 & 10.50 & 10.50 & 10.43 \\
MaxLogits & 10.00 & 10.00 & 8.00 & 9.33 & 11.00 & 10.00 & 10.17 & 9.25 & 7.50 & 8.33 & 9.38 \\
EmbeddingMLP & 10.00 & 15.50 & 10.00 & 11.83 & 9.00 & 13.50 & 11.42 & 9.25 & 9.67 & 11.50 & 10.72 \\
\cellcolor{Highlight}ProbeDirichlet 
& \cellcolor{Highlight}13.50 & \cellcolor{Highlight}\textbf{21.00} & \cellcolor{Highlight}\textbf{21.00} & \cellcolor{Highlight}\textbf{18.50}
& \cellcolor{Highlight}\textbf{14.50} & \cellcolor{Highlight}\textbf{21.00} & \cellcolor{Highlight}15.75
& \cellcolor{Highlight}11.50 & \cellcolor{Highlight}\textbf{14.83} & \cellcolor{Highlight}\textbf{14.83} & \cellcolor{Highlight}\textbf{15.40} \\
\bottomrule
\end{tabular}
}
\label{tab:scence}
\end{table*}

\section{Experiments}
\subsection{Experimental setup}

\paragraph{Benchmarks.} 
\label{sec:exp_benchmarks}

We evaluate routers on six representative benchmarks.  
For training and in-domain evaluation, we use \textit{Alpaca}~\citep{alpaca} (general tasks), \textit{MMLU}~\citep{hendrycks2021measuring} (knowledge), and \textit{Big-Math}~\citep{albalak2025bigmath} (math).  
For out-of-domain evaluation, we use \textit{Magpie}~\citep{xu2024magpie} (general tasks), \textit{MMLU Pro}~\citep{wang2024mmlupro} (knowledge, covering STEM, Humanities, Social Sciences, and Others), and \textit{MATH}~\citep{hendrycks2021math} (math).  
The benchmark design is guided by three principles.  
Task coverage is ensured by including general, knowledge, and math domains.  
The difficulty gradient is reflected in the progression from simpler benchmarks such as \textit{Alpaca}, \textit{Magpie}, to more challenging ones like \textit{MMLU}, \textit{Big-Math}, and \textit{MATH}. 
Detailed data preparation and specific evaluation protocols are provided in Appendix~\ref{app:datasets}.

For model selection, we use \textit{GPT-5} as the large model and \textit{Llama-3.1-8B-Instruct} as the small model for evaluating router performance.  

\paragraph{Baselines.}  
We compare our hidden-state approach against three alternative signal modalities: 
\textbf{(1) Verbose-based.}  
Routers that depend on auxiliary generations, such as self-evaluation 
\citep{kadavath2022languagemodelsmostlyknow,ding2025bestroute} 
or semantic entropy \citep{kuhn2023semanticuncertaintylinguisticinvariances,zhang2025leveraginguncertaintyestimationefficient}, 
which are informative but incur prompt sensitivity.
\textbf{(2) Logit-based.}  
Routers that only use the final-layer logits, such as entropy \citep{su2025cprouteruncertaintyawarerouterllm}, margin \citep{rez2024optimising}. These are efficient but brittle across domains.
\textbf{(3) Embedding-based.}
These routers use fixed pretrained encoders with lightweight classifiers for semantic representations~\citep{feng2025fusionfactory}. 
With comparable classifier sizes, this enables direct comparison of different routing signals.
By categorizing baselines via their signal sources, we can facilitate a systematic comparison of different signal modalities.

\paragraph{Training Setup.} 
For all probe-based methods, we use a lightweight linear model with input dimension 4096, corresponding to the small model's hidden state size.
All models are trained with a fixed random seed. Training proceeds for 50 epochs with a learning rate of $1\times10^{-4}$.  
The training data consists of 12K examples, combining MMLU, Big Math, and Alpaca with 4K samples each.

\subsection{Main Results}
\paragraph{Router Ability.}
Table \ref{tab:perf_char} reports the overall routing accuracy across 
multiple benchmarks. Our hidden-state–based strategies achieve 16.68\% 
relative improvement over the best baseline in both in-domain and 
out-of-distribution scenarios. Within our approaches, ProbeDirichlet 
achieves marginally higher performance than ProbeMean through learned 
distributional layer weights. However, both variants perform competitively, indicating that strong 
results stem primarily from the hidden-state signals themselves rather than 
the aggregation mechanism.
These results demonstrate that signal provenance is crucial: internal representations encode task-model interactions that external features cannot capture.

\paragraph{Scenario Alignment.}
Our framework enables flexible scenario definition based on deployment needs. 
Table~\ref{tab:scence} demonstrates router performance across three  
scenarios: cost-sensitive (LPM at 25-30\% call rate), balanced (MPM), and 
accuracy-critical (HCR at 85-95\% relative performance).

Probe-based methods outperform all baselines, especially in accuracy-critical scenarios.  
In cost-sensitive and balanced regimes, performance differences remain modest because routers only need to escalate obviously difficult queries—a task most signal types handle adequately. However, accuracy-critical scenarios require precise identification of boundary cases where small models approach but do not meet requirements. Here, probe-based methods achieve 18.86\% relative improvement, demonstrating that fine-grained difficulty discrimination requires richer internal signals.

\subsection{Ablation Study}

Table~\ref{tab:ablation_agg} compares three probe aggregation strategies: 
\textit{Final} uses only the last layer, \textit{Mean} uniformly averages 
all layers, and \textit{Dirichlet} is our proposed method. Results 
show that our method achieves the best AUROC 
across all datasets.

\begin{table}[h]
    \centering
    \footnotesize
    \caption{AUROC (\%) of probe aggregation methods.}
    \label{tab:ablation_agg}
    \begin{tabular}{lcccc}
        \toprule
        & \textbf{Alpaca} & \textbf{BigMath} & \textbf{MMLU} & \textbf{Average} \\
        \midrule
        Final Layer  & 61.97  & 50.33  &  49.45 & 53.91  \\
        Mean Pool    & 71.34 & 65.69 & 67.10 & 68.04 \\
        Dirichlet    & 72.02 & 66.18 & 67.88 & 68.70 \\
        \bottomrule
    \end{tabular}
\end{table}
Dirichlet achieves the best performance, and both aggregation methods significantly outperform the Final Layer baseline, confirming that cross-layer aggregation better captures task difficulty.

\section{Analysis}

\paragraph{Internal Hidden States Matter.}
To isolate the impact of signal source from model architecture, we compare 
three input representations using identical linear models: 
Longformer embeddings, LLM embeddings, and LLM hidden states. 
\begin{table}[h]
\centering
\footnotesize
\caption{Performance comparison across different input representations.}
\label{tab:signal_source}
\begin{tabular}{lcccc}
\toprule
\textbf{Source} & \textbf{Alpaca} & \textbf{BigMath} & \textbf{Magpie} & \textbf{MATH} \\
\midrule
Longformer & 61.95 & 43.10 & 66.19 & 42.52 \\
LLM Emb.   & 62.47 & 56.21 & 66.22 & 58.82 \\
LLM Hidden & 71.34 & 62.39 & 74.31 & 67.73 \\
\bottomrule
\end{tabular}
\end{table}

Table~\ref{tab:signal_source} shows that LLM hidden states significantly outperform embedding-based methods, with particularly strong gains on mathematical reasoning tasks. This indicates that intermediate representations preserve richer hierarchical information. While the embedding layer only provides raw lexical representations, hidden states encode multi-scale features from low-level syntax to high-level semantics through  Transformer layers. Mathematical reasoning depends on multi-level signals, including symbolic correctness and logical coherence, while instruction-following tasks rely mainly on surface-level semantic matching. Although LLM embeddings show a slight advantage over Longformer, likely due to vocabulary alignment, the improvement remains substantially smaller than that obtained from intermediate-layer representations. These results suggest that quality prediction should prioritize internal hierarchical representations rather than relying solely on input-layer features or external encoders.

\paragraph{Impact of Probe Architecture.}
To verify that lightweight architectures suffice, we compare a linear probe 
with a two-layer MLP under the mixed-dataset training setting.

\begin{figure}[h]
    \centering
    \begin{subfigure}[t]{0.49\linewidth}
        \centering
        \includegraphics[width=\linewidth]{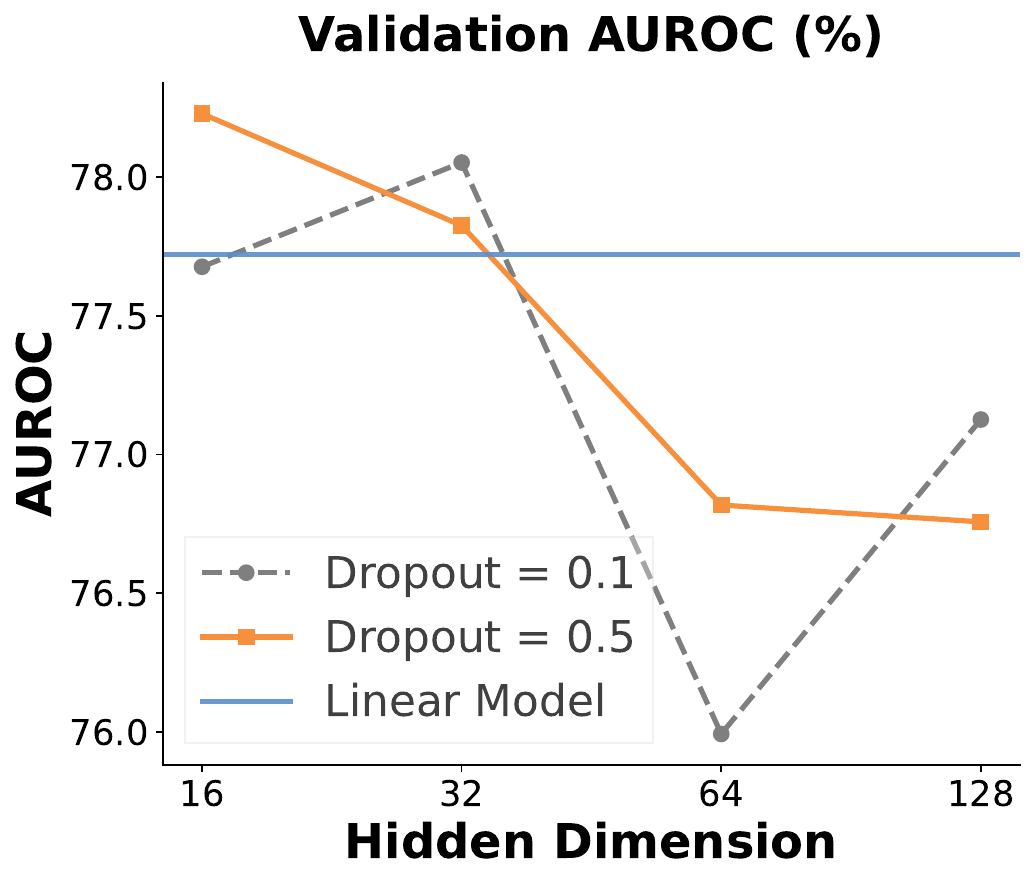}
        \label{fig:layer2_auroc}
    \end{subfigure}
    \hfill
    \begin{subfigure}[t]{0.49\linewidth}
        \centering
        \includegraphics[width=\linewidth]{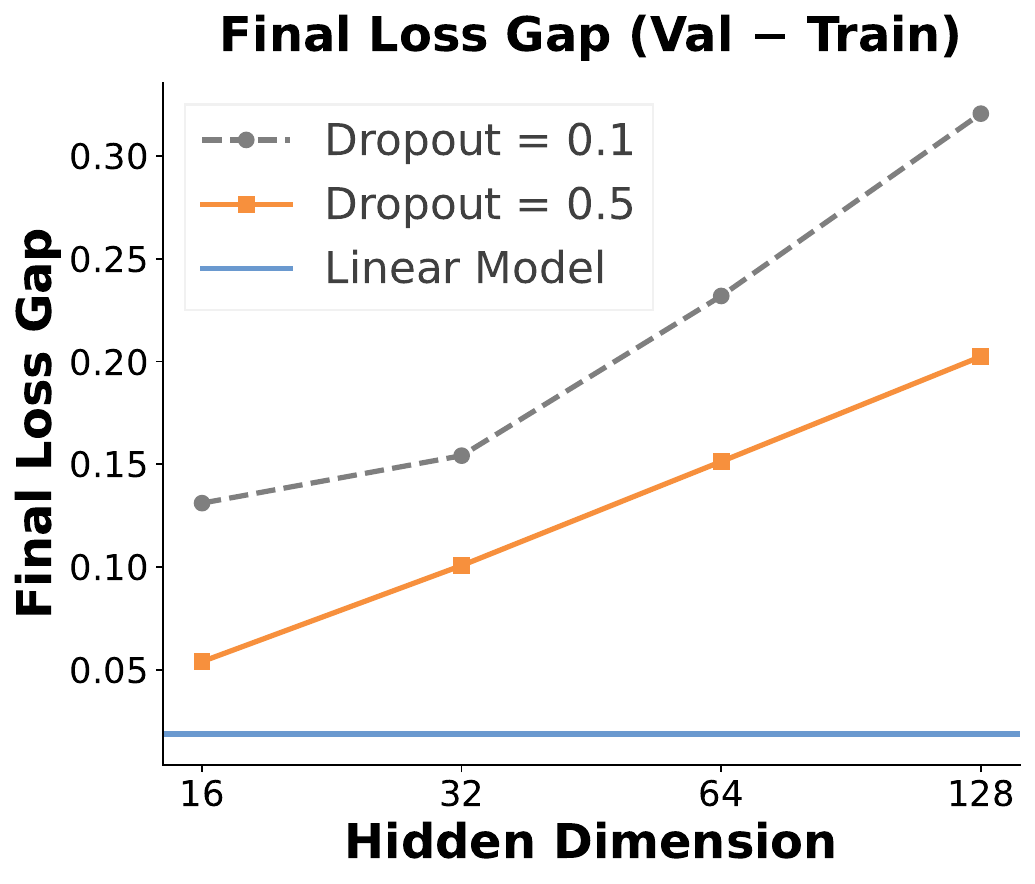}
        \label{fig:layer2_loss_gap}
    \end{subfigure}

  \caption{Effect of probe complexity on performance and generalization.
The \textbf{horizontal line} represents the \textbf{Linear Probe baseline}, serving as a constant reference independent of the hidden dimension axis.}
    \label{fig:layer2_hidden_dim_ablation}
\end{figure}

Figure~\ref{fig:layer2_hidden_dim_ablation} compares one-hidden-layer MLPs 
with the linear baseline (dashed line). Introducing hidden layers provides 
almost no performance benefit but substantially increases overfitting, as 
evidenced by widening train-validation loss gaps. These results indicate 
that increasing model complexity is unnecessary for effective routing: a 
linear probe already achieves comparable or better performance, and 
introducing non-linearity or extra layers does not provide additional benefit.
\paragraph{Scaling Provides Diminishing Returns.}
We examine whether increasing training data improves probe performance by 
training on varying amounts of data from individual datasets. 
We split each dataset into fixed train/test sets, 
train probes on different data scales, and evaluate 
all models on their respective held-out test sets.

\begin{figure}[H]
    \centering
    \includegraphics[width=1\linewidth]{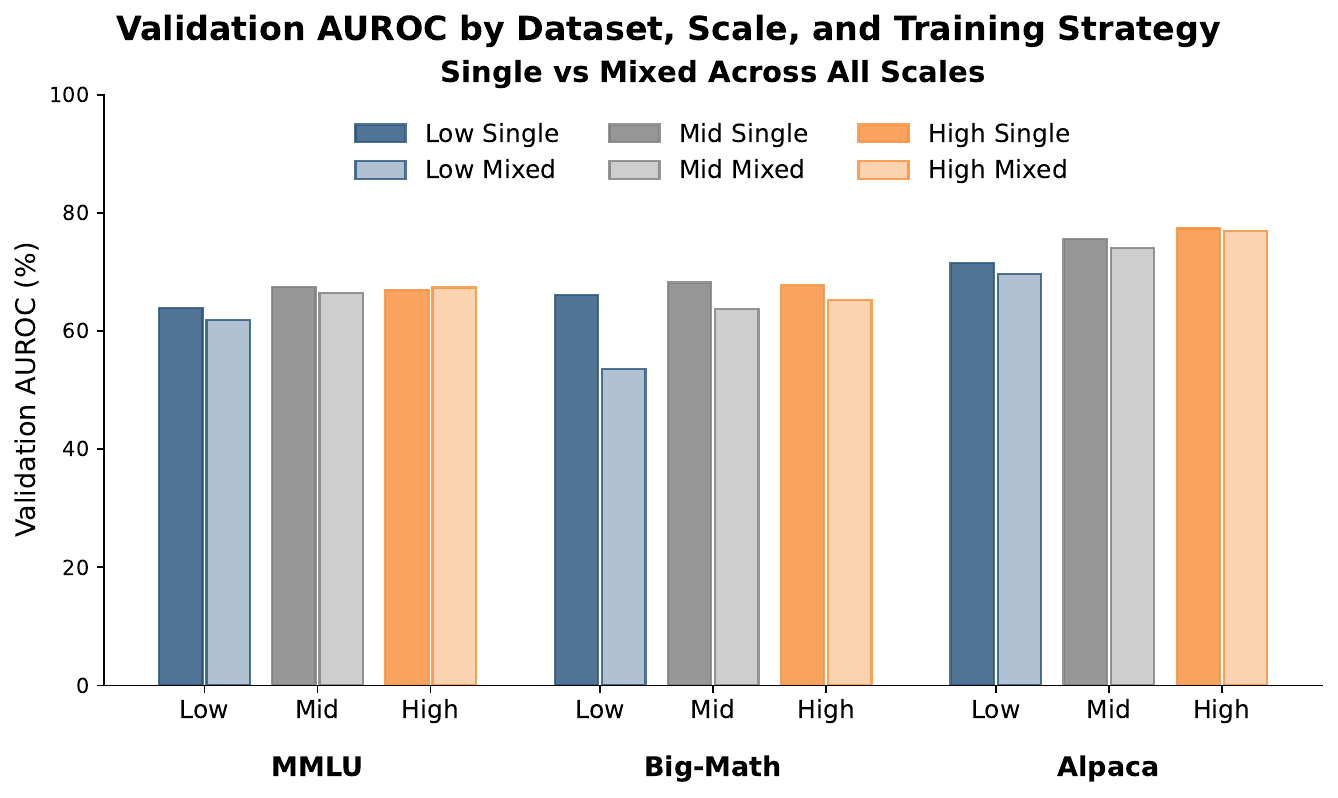}
    \caption{Validation AUROC (\%) across training scales for single-dataset and mixed-dataset probes. Low/Mid/High denote 1K/4K/8K samples per dataset for single-dataset training, and 3K/12K/24K total samples for mixed training.}
    \label{fig:probe_scaling}
\end{figure}

Figure~\ref{fig:probe_scaling} shows that 1K samples are insufficient, 
with performance substantially lower across all settings. However, scaling 
from 4K to 8K yields minimal gains, indicating that probes 
saturate quickly once they capture sufficient signal. The mixed-corpus probe 
matches single-dataset performance at high scale, demonstrating that data 
diversity compensates for domain-specific concentration.Given these results, we ask whether adding domains creates interference or instead yields additive gains.

\begin{table*}
\centering
\caption{Comparison of EmbeddingMLP and ProbeDirichlet performance across model families and scales.}
\label{tab:cross_family}
\small
    \resizebox{\textwidth}{!}{
    \begin{tabular}{ll|cccc|ccccccc}
    \toprule
    \multirow{2}{*}{\textbf{Model}}
    & \multirow{2}{*}{\textbf{Method}}
    & \multicolumn{4}{c|}{\textbf{In-Domain}}
    & \multicolumn{7}{c}{\textbf{Out-of-Domain}} \\
    \cmidrule(lr){3-6} \cmidrule(lr){7-13}
    & & Alpaca & BigMath & MMLU & \textbf{AVG}
    & Magpie & MATH & STEM &  Human. & Social Sci. & Others & \textbf{AVG} \\
\midrule
\multirow{2}{*}{Llama-3.1-8B-Instruct}
& EmbeddingMLP & 67.31 & 56.18 & 54.89 & 59.46 & 68.97 & 56.97 & 52.97 & 53.77 & 48.16 & 50.45 & 55.22 \\
& ProbeDirichlet & 72.02 & 66.18 & 67.88 & \textbf{68.70} & 74.08 & 73.90 & 65.32 & 57.84 & 58.82 & 62.77 & \textbf{65.46} \\
\midrule
\multirow{2}{*}{Qwen2.5-0.5B-Instruct}
& EmbeddingMLP & 59.52 & 60.50 & 53.53 & 57.85 & 73.18 & 55.40 & 43.38 & 52.97 & 51.72 & 51.19 & 54.64 \\
& ProbeDirichlet & 65.71 & 67.87 & 60.96 & \textbf{64.84} & 74.40 & 61.78 & 60.69 & 50.50 & 52.70 & 56.13 & \textbf{59.36} \\

\midrule
\multirow{2}{*}{Qwen2.5-3B-Instruct}
& EmbeddingMLP & 59.51 & 62.14 & 52.53 & 58.06 & 76.38 & 55.35 & 45.55 & 48.32 & 54.61 & 46.80 & 54.50 \\
& ProbeDirichlet & 67.90 & 70.72 & 68.88 & \textbf{69.17} & 82.99 & 77.18 & 66.43 & 55.83 & 58.78 & 61.62 & \textbf{67.14} \\

\midrule
\multirow{2}{*}{Qwen2.5-7B-Instruct}
& EmbeddingMLP & 61.36 & 59.65 & 55.66 & 58.89 & 76.56 & 55.00 & 46.23 & 48.53 & 56.04 & 49.99 & 55.39 \\
& ProbeDirichlet & 69.60 & 78.03 & 73.17 & \textbf{73.60} & 81.41 & 77.77 & 64.85 & 54.61 & 58.17 & 60.51 & \textbf{66.22} \\

\bottomrule
\end{tabular}
}
\end{table*}

\paragraph{Data Diversity Yields Additive Gains Without Interference.}

We train on progressively larger data mixtures. As shown in Table~\ref{tab:probe_data_generation}, the results show striking additive gains: existing 
performance is preserved (Alpaca: 71.85→71.96) while new domains contribute 
independently (BigMath: 49.19→66.18; MMLU: 49.35→67.88). This pattern explains 
why lightweight probes suffice. If domains conflicted, adding BigMath would 
degrade Alpaca. However, we observe no such interference; domains coexist 
harmoniously, suggesting hidden states encode a shared notion of difficulty 
that simple models can generalize across diverse tasks. Data diversity is additive, not competitive; diverse training improves robustness while preserving specialist capabilities.

\begin{table}[H]
\centering
\caption{Generalization Behavior under Different Dataset Compositions}
\small
\setlength{\tabcolsep}{2pt}
\renewcommand{\arraystretch}{1.15}
\begin{tabular}{lccc}
\toprule
\textbf{Benchmark} &
\textbf{Alpaca} &
\textbf{Alpaca + BigMath} &
\textbf{Mixed Training} \\
\midrule
\multicolumn{4}{l}{\textit{In-domain}} \\
Alpaca   & 71.85 & 71.63 & \textbf{72.02} \\
BigMath  & 49.19 & \textbf{66.49} & 66.18 \\
MMLU     & 49.35 & 51.06 & \textbf{67.88} \\
\midrule
\multicolumn{4}{l}{\textit{Out-of-domain}} \\
Magpie   & 72.80 & \textbf{74.32} & 74.08 \\

MATH     & 57.97 & 72.64 & \textbf{73.90} \\
MMLU-Pro & 48.41 & 49.62 & \textbf{61.19} \\
\bottomrule
\label{tab:probe_data_generation}
\end{tabular}
\end{table}

\paragraph{Generalization Across Model Families.}
To verify that our approach is not specific to Llama-3.1, we train 
and evaluate ProbeDirichlet on the Qwen2.5-Instruct family.

Table~\ref{tab:cross_family} demonstrates consistent effectiveness 
across architectures. ProbeDirichlet significantly outperforms the 
EmbeddingMLP baseline across all models, with an average improvement 
of 10.5\% on in-domain tasks and 9.6\% on out-of-domain tasks.
 Within the Qwen family, we observe distinct scaling patterns. While 
in-domain accuracy improves monotonically with model size, out-of-domain 
performance 
varies by task type: mathematical reasoning plateaus at larger scales, 
instruction-following tasks show non-linear scaling effects, while 
knowledge-based tasks remain relatively stable. Taken together, the consistent performance across architectures and varying scaling patterns across tasks demonstrate the broad applicability of our 
routing approach.

\begin{figure}[t]
    \centering
    \includegraphics[width=1\linewidth]{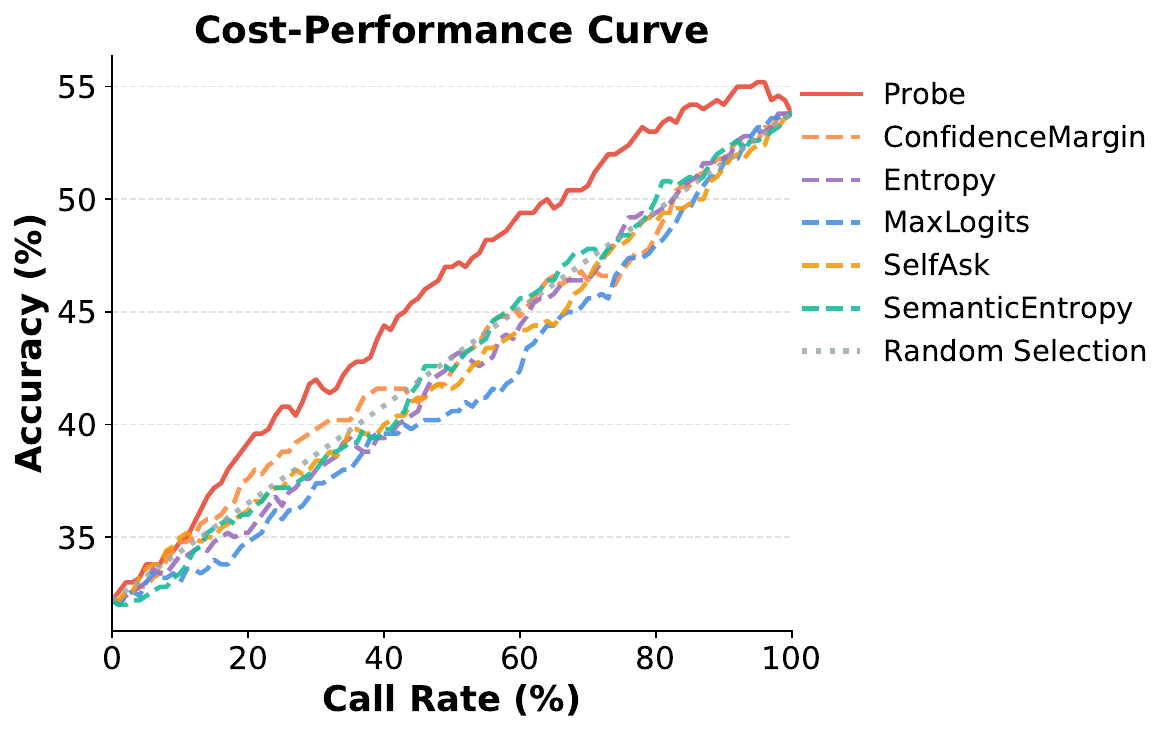}
    \caption{Cost-Performance curve under the agent-based inference scenario on HotpotQA.}
    \label{fig:agent_callrate_curve}
\end{figure}

\paragraph{Agent-based Inference Scenario.}
Beyond model collaboration, our router generalizes to agent-based inference, 
deciding when tool-augmented reasoning is needed.
We evaluate it in HotpotQA, which requires multi-hop reasoning and iterative evidence retrieval.
Figure~\ref{fig:agent_callrate_curve} demonstrates robust generalization 
to agent scenarios. Our router shows a clear advantage across 
the entire cost-accuracy frontier.

\section{Conclusion}
We present a principled evaluation framework that disentangles intrinsic routing ability from scenario-specific requirements across three dimensions: router ability (AUROC), scenario alignment (LPM, MPM, HCR), and cross-domain robustness, enabling fair comparison under diverse deployment constraints. We further introduce a hidden-state router trained across multiple domains, which consistently outperforms baselines on standard benchmarks and agentic workflows. Our analysis shows that robustness is driven by training data diversity rather than architectural complexity, offering practical guidance for collaborative LLM systems.


\section*{Limitations}

Our routing framework assumes the large model's capability exceeds the small 
model's; however, both models may perform similarly or converge on the same 
incorrect answer in certain domains (Appendix~\ref{app:case_study}), limiting 
routing effectiveness. Our experiments focus on a single small-large model pair 
and report single-run results due to computational constraints; broader validation 
across diverse architectures, multiple seeds, and more complex OOD conditions 
would further strengthen the conclusions.

\bibliography{custom}

\begin{thebibliography}{54}
\providecommand{\natexlab}[1]{#1}

\bibitem[{Aggarwal et~al.(2024)Aggarwal, Madaan, Anand, Potharaju, Mishra, Zhou, Gupta, Rajagopal, Kappaganthu, Yang, Upadhyay, Faruqui, and Mausam}]{aggarwalAutoMixAutomaticallyMixing2024}
Pranjal Aggarwal, Aman Madaan, Ankit Anand, Srividya~Pranavi Potharaju, Swaroop Mishra, Pei Zhou, Aditya Gupta, Dheeraj Rajagopal, Karthik Kappaganthu, Yiming Yang, Shyam Upadhyay, Manaal Faruqui, and Mausam. 2024.
\newblock \href {https://doi.org/10.48550/arXiv.2310.12963} {{{AutoMix}}: {{Automatically Mixing Language Models}}}.

\bibitem[{Albalak et~al.(2025)Albalak, Phung, Lile, Rafailov, Gandhi, Castricato, Singh, Blagden, Xiang, Mahan, and Haber}]{albalak2025bigmath}
Alon Albalak, Duy Phung, Nathan Lile, Rafael Rafailov, Kanishk Gandhi, Louis Castricato, Anikait Singh, Chase Blagden, Violet Xiang, Dakota Mahan, and Nick Haber. 2025.
\newblock \href {https://arxiv.org/abs/2502.17387} {Big-math: A large-scale, high-quality math dataset for reinforcement learning in language models}.
\newblock \emph{Preprint}, arXiv:2502.17387.

\bibitem[{Barrak et~al.(2025)Barrak, Fourati, Olchawa, Ksontini, and Zoghlami}]{barrak2025cargoframeworkconfidenceawarerouting}
Amine Barrak, Yosr Fourati, Michael Olchawa, Emna Ksontini, and Khalil Zoghlami. 2025.
\newblock \href {https://arxiv.org/abs/2509.14899} {Cargo: A framework for confidence-aware routing of large language models}.
\newblock \emph{Preprint}, arXiv:2509.14899.

\bibitem[{Busch et~al.(2025)Busch, Hoffmann, Rueger, van Dijk, Kader, Ortiz-Prado, Makowski, Saba, Hadamitzky, Kather, Truhn, Cuocolo, Adams, and Bressem}]{buschCurrentApplicationsChallenges2025c}
Felix Busch, Lena Hoffmann, Christopher Rueger, Elon H.~C. van Dijk, Rawen Kader, Esteban Ortiz-Prado, Marcus~R. Makowski, Luca Saba, Martin Hadamitzky, Jakob~Nikolas Kather, Daniel Truhn, Renato Cuocolo, Lisa~C. Adams, and Keno~K. Bressem. 2025.
\newblock \href {https://doi.org/10.1038/s43856-024-00717-2} {Current applications and challenges in large language models for patient care: A systematic review}.
\newblock \emph{Communications Medicine}, 5(1):26.

\bibitem[{Cai et~al.(2024)Cai, Li, Geng, Peng, Lee, Chen, and Dao}]{cai2024medusasimplellminference}
Tianle Cai, Yuhong Li, Zhengyang Geng, Hongwu Peng, Jason~D. Lee, Deming Chen, and Tri Dao. 2024.
\newblock \href {https://arxiv.org/abs/2401.10774} {Medusa: Simple llm inference acceleration framework with multiple decoding heads}.
\newblock \emph{Preprint}, arXiv:2401.10774.

\bibitem[{Chen et~al.(2024{\natexlab{a}})Chen, Liu, Chen, Gu, Wu, Tao, Fu, and Ye}]{chen2024inside}
Chao Chen, Kai Liu, Ze~Chen, Yi~Gu, Yue Wu, Mingyuan Tao, Zhihang Fu, and Jieping Ye. 2024{\natexlab{a}}.
\newblock \href {https://openreview.net/forum?id=Zj12nzlQbz} {{INSIDE}: {LLM}s' internal states retain the power of hallucination detection}.
\newblock In \emph{The Twelfth International Conference on Learning Representations}.

\bibitem[{Chen et~al.(2023)Chen, Borgeaud, Irving, Lespiau, Sifre, and Jumper}]{chen2023acceleratinglargelanguagemodel}
Charlie Chen, Sebastian Borgeaud, Geoffrey Irving, Jean-Baptiste Lespiau, Laurent Sifre, and John Jumper. 2023.
\newblock \href {https://arxiv.org/abs/2302.01318} {Accelerating large language model decoding with speculative sampling}.
\newblock \emph{Preprint}, arXiv:2302.01318.

\bibitem[{Chen et~al.(2024{\natexlab{b}})Chen, Zaharia, and Zou}]{chen2024frugalgpt}
Lingjiao Chen, Matei Zaharia, and James Zou. 2024{\natexlab{b}}.
\newblock \href {https://openreview.net/forum?id=cSimKw5p6R} {Frugal{GPT}: How to use large language models while reducing cost and improving performance}.
\newblock \emph{Transactions on Machine Learning Research}.

\bibitem[{Chen et~al.(2024{\natexlab{c}})Chen, Jiang, Lin, Kwok, and Zhang}]{chen2024routerdc}
Shuhao Chen, Weisen Jiang, Baijiong Lin, James Kwok, and Yu~Zhang. 2024{\natexlab{c}}.
\newblock \href {https://openreview.net/forum?id=7RQvjayHrM} {Router{DC}: Query-based router by dual contrastive learning for assembling large language models}.
\newblock In \emph{The Thirty-eighth Annual Conference on Neural Information Processing Systems}.

\bibitem[{Ding et~al.(2024{\natexlab{a}})Ding, Mallick, Wang, Sim, Mukherjee, Ruhle, Lakshmanan, and Awadallah}]{dingHybridLLMCostEfficient2024}
Dujian Ding, Ankur Mallick, Chi Wang, Robert Sim, Subhabrata Mukherjee, Victor Ruhle, Laks V.~S. Lakshmanan, and Ahmed~Hassan Awadallah. 2024{\natexlab{a}}.
\newblock \href {https://doi.org/10.48550/arXiv.2404.14618} {Hybrid {{LLM}}: {{Cost-Efficient}} and {{Quality-Aware Query Routing}}}.
\newblock \emph{Preprint}, arXiv:2404.14618.

\bibitem[{Ding et~al.(2024{\natexlab{b}})Ding, Mallick, Wang, Sim, Mukherjee, Ruhle, Lakshmanan, and Awadallah}]{ding2024hybridllmcostefficientqualityaware}
Dujian Ding, Ankur Mallick, Chi Wang, Robert Sim, Subhabrata Mukherjee, Victor Ruhle, Laks V.~S. Lakshmanan, and Ahmed~Hassan Awadallah. 2024{\natexlab{b}}.
\newblock \href {https://arxiv.org/abs/2404.14618} {Hybrid llm: Cost-efficient and quality-aware query routing}.
\newblock \emph{Preprint}, arXiv:2404.14618.

\bibitem[{Ding et~al.(2025)Ding, Mallick, Zhang, Wang, Madrigal, Garcia, Xia, Lakshmanan, Wu, and R{\"u}hle}]{ding2025bestroute}
Dujian Ding, Ankur Mallick, Shaokun Zhang, Chi Wang, Daniel Madrigal, Mirian Del Carmen~Hipolito Garcia, Menglin Xia, Laks V.~S. Lakshmanan, Qingyun Wu, and Victor R{\"u}hle. 2025.
\newblock \href {https://openreview.net/forum?id=tFBIbCVXkG} {{BEST}-route: Adaptive {LLM} routing with test-time optimal compute}.
\newblock In \emph{Forty-second International Conference on Machine Learning}.

\bibitem[{Dong et~al.(2025)Dong, Zhou, Liu, Yang, and Lu}]{dong2025emergent}
Zhichen Dong, Zhanhui Zhou, Zhixuan Liu, Chao Yang, and Chaochao Lu. 2025.
\newblock \href {https://openreview.net/forum?id=Ce79P8ULPY} {Emergent response planning in {LLMs}}.
\newblock In \emph{Proceedings of the 42nd International Conference on Machine Learning}. PMLR.

\bibitem[{Duan et~al.(2024)Duan, Cheng, Wang, Zavalny, Wang, Xu, Kailkhura, and Xu}]{duan2024shifting}
Jinhao Duan, Hao Cheng, Shiqi Wang, Alex Zavalny, Chenan Wang, Renjing Xu, Bhavya Kailkhura, and Kaidi Xu. 2024.
\newblock \href {https://openreview.net/forum?id=yZJapMWdHZ} {Shifting attention to relevance: Towards the uncertainty estimation of large language models}.

\bibitem[{Fadeeva et~al.(2024)Fadeeva, Rubashevskii, Shelmanov, Petrakov, Li, Mubarak, Tsymbalov, Kuzmin, Panchenko, Baldwin, Nakov, and Panov}]{fadeeva-etal-2024-fact}
Ekaterina Fadeeva, Aleksandr Rubashevskii, Artem Shelmanov, Sergey Petrakov, Haonan Li, Hamdy Mubarak, Evgenii Tsymbalov, Gleb Kuzmin, Alexander Panchenko, Timothy Baldwin, Preslav Nakov, and Maxim Panov. 2024.
\newblock \href {https://doi.org/10.18653/v1/2024.findings-acl.558} {Fact-checking the output of large language models via token-level uncertainty quantification}.
\newblock In \emph{Findings of the Association for Computational Linguistics: ACL 2024}, pages 9367--9385, Bangkok, Thailand. Association for Computational Linguistics.

\bibitem[{Feng et~al.(2025)Feng, Zhang, Lei, Han, Patwary, Shoeybi, Catanzaro, and You}]{feng2025fusionfactory}
Tao Feng, Haozhen Zhang, Zijie Lei, Pengrui Han, Mostofa Patwary, Mohammad Shoeybi, Bryan Catanzaro, and Jiaxuan You. 2025.
\newblock \href {https://arxiv.org/abs/2507.10540} {Fusionfactory: Fusing {LLM} capabilities with multi-{LLM} log data}.
\newblock \emph{Preprint}, arXiv:2507.10540.

\bibitem[{Fomicheva et~al.(2020)Fomicheva, Sun, Yankovskaya, Blain, Guzm{\'a}n, Fishel, Aletras, Chaudhary, and Specia}]{fomicheva-etal-2020-unsupervised}
Marina Fomicheva, Shuo Sun, Lisa Yankovskaya, Fr{\'e}d{\'e}ric Blain, Francisco Guzm{\'a}n, Mark Fishel, Nikolaos Aletras, Vishrav Chaudhary, and Lucia Specia. 2020.
\newblock \href {https://doi.org/10.1162/tacl_a_00330} {Unsupervised quality estimation for neural machine translation}.
\newblock \emph{Transactions of the Association for Computational Linguistics}, 8:539--555.

\bibitem[{Guha et~al.(2024)Guha, Chen, Chow, Khare, and Re}]{guha2024smoothie}
Neel Guha, Mayee~F Chen, Trevor Chow, Ishan~S. Khare, and Christopher Re. 2024.
\newblock \href {https://openreview.net/forum?id=pPSWHsgqRp} {Smoothie: Label free language model routing}.
\newblock In \emph{The Thirty-eighth Annual Conference on Neural Information Processing Systems}.

\bibitem[{Guo et~al.(2017)Guo, Pleiss, Sun, and Weinberger}]{guo2017calibration}
Chuan Guo, Geoff Pleiss, Yu~Sun, and Kilian~Q. Weinberger. 2017.
\newblock \href {https://proceedings.mlr.press/v70/guo17a.html} {On calibration of modern neural networks}.
\newblock In \emph{Proceedings of the 34th International Conference on Machine Learning}, volume~70 of \emph{Proceedings of Machine Learning Research}, pages 1321--1330. PMLR.

\bibitem[{Gupta et~al.(2024)Gupta, Narasimhan, Jitkrittum, Rawat, Menon, and Kumar}]{gupta2024language}
Neha Gupta, Harikrishna Narasimhan, Wittawat Jitkrittum, Ankit~Singh Rawat, Aditya~Krishna Menon, and Sanjiv Kumar. 2024.
\newblock \href {https://openreview.net/forum?id=KgaBScZ4VI} {Language model cascades: Token-level uncertainty and beyond}.
\newblock In \emph{The Twelfth International Conference on Learning Representations}.

\bibitem[{Hendrycks et~al.(2021{\natexlab{a}})Hendrycks, Burns, Basart, Zou, Mazeika, Song, and Steinhardt}]{hendrycks2021measuring}
Dan Hendrycks, Collin Burns, Steven Basart, Andy Zou, Mantas Mazeika, Dawn Song, and Jacob Steinhardt. 2021{\natexlab{a}}.
\newblock \href {https://arxiv.org/abs/2009.03300} {Measuring massive multitask language understanding}.
\newblock In \emph{International Conference on Learning Representations (ICLR)}.

\bibitem[{Hendrycks et~al.(2021{\natexlab{b}})Hendrycks, Burns, Kadavath, Arora, Basart, Tang, Song, and Steinhardt}]{hendrycks2021math}
Dan Hendrycks, Collin Burns, Saurav Kadavath, Akul Arora, Steven Basart, Eric Tang, Dawn Song, and Jacob Steinhardt. 2021{\natexlab{b}}.
\newblock \href {https://arxiv.org/abs/2103.03874} {Measuring mathematical problem solving with the math dataset}.
\newblock In \emph{Advances in Neural Information Processing Systems (NeurIPS)}.

\bibitem[{Hu et~al.(2024)Hu, Bieker, Li, Jiang, Keigwin, Ranganath, Keutzer, and Upadhyay}]{hu2024routerbench}
Qitian~Jason Hu, Jacob Bieker, Xiuyu Li, Nan Jiang, Benjamin Keigwin, Gaurav Ranganath, Kurt Keutzer, and Shriyash~Kaustubh Upadhyay. 2024.
\newblock \href {https://openreview.net/forum?id=IVXmV8Uxwh} {Routerbench: A benchmark for multi-{LLM} routing system}.
\newblock In \emph{Agentic Markets Workshop at ICML 2024}.

\bibitem[{Huang et~al.(2025)Huang, Ling, Lin, Chen, Zhong, Wu, and Lin}]{hu2025routereval}
Zhongzhan Huang, Guoming Ling, Yupei Lin, Yandong Chen, Shanshan Zhong, Hefeng Wu, and Liang Lin. 2025.
\newblock \href {https://doi.org/10.18653/v1/2025.findings-emnlp.208} {Routereval: A comprehensive benchmark for routing llms to explore model-level scaling up in llms}.
\newblock In \emph{Findings of the Association for Computational Linguistics: EMNLP 2025}. Association for Computational Linguistics.

\bibitem[{Kadavath et~al.(2022)Kadavath, Conerly, Askell, Henighan, Drain, Perez, Schiefer, Hatfield-Dodds, DasSarma, Tran-Johnson, Johnston, El-Showk, Jones, Elhage, Hume, Chen, Bai, Bowman, Fort, Ganguli, Hernandez, Jacobson, Kernion, Kravec, Lovitt, Ndousse, Olsson, Ringer, Amodei, Brown, Clark, Joseph, Mann, McCandlish, Olah, and Kaplan}]{kadavath2022languagemodelsmostlyknow}
Saurav Kadavath, Tom Conerly, Amanda Askell, Tom Henighan, Dawn Drain, Ethan Perez, Nicholas Schiefer, Zac Hatfield-Dodds, Nova DasSarma, Eli Tran-Johnson, Scott Johnston, Sheer El-Showk, Andy Jones, Nelson Elhage, Tristan Hume, Anna Chen, Yuntao Bai, Sam Bowman, Stanislav Fort, and 17 others. 2022.
\newblock \href {https://arxiv.org/abs/2207.05221} {Language models (mostly) know what they know}.
\newblock \emph{Preprint}, arXiv:2207.05221.

\bibitem[{Kassem et~al.(2025)Kassem, Schölkopf, and Jin}]{kassem2025robustrouterllmsanalysisfragility}
Aly~M. Kassem, Bernhard Schölkopf, and Zhijing Jin. 2025.
\newblock \href {https://arxiv.org/abs/2504.07113} {How robust are router-llms? analysis of the fragility of llm routing capabilities}.
\newblock \emph{Preprint}, arXiv:2504.07113.

\bibitem[{Kuhn et~al.(2023{\natexlab{a}})Kuhn, Gal, and Farquhar}]{kuhn2023semantic}
Lorenz Kuhn, Yarin Gal, and Sebastian Farquhar. 2023{\natexlab{a}}.
\newblock \href {https://openreview.net/forum?id=VD-AYtP0dve} {Semantic uncertainty: Linguistic invariances for uncertainty estimation in natural language generation}.
\newblock In \emph{The Eleventh International Conference on Learning Representations}.

\bibitem[{Kuhn et~al.(2023{\natexlab{b}})Kuhn, Gal, and Farquhar}]{kuhn2023semanticuncertaintylinguisticinvariances}
Lorenz Kuhn, Yarin Gal, and Sebastian Farquhar. 2023{\natexlab{b}}.
\newblock \href {https://arxiv.org/abs/2302.09664} {Semantic uncertainty: Linguistic invariances for uncertainty estimation in natural language generation}.
\newblock \emph{Preprint}, arXiv:2302.09664.

\bibitem[{Li et~al.(2023)Li, Hammoud, Itani, Khizbullin, and Ghanem}]{li2023camel}
Guohao Li, Hasan Abed Al~Kader Hammoud, Hani Itani, Dmitrii Khizbullin, and Bernard Ghanem. 2023.
\newblock \href {https://openreview.net/forum?id=3IyL2XWDkG} {{CAMEL}: Communicative agents for ''mind'' exploration of large language model society}.
\newblock In \emph{Thirty-seventh Conference on Neural Information Processing Systems}.

\bibitem[{Li(2025)}]{li2025llm}
Yang Li. 2025.
\newblock \href {https://openreview.net/forum?id=rEqETC88RY} {{LLM} bandit: Cost-efficient {LLM} generation via preference-conditioned dynamic routing}.

\bibitem[{Li et~al.(2024)Li, Wei, Zhang, and Zhang}]{li2024eagle2fasterinferencelanguage}
Yuhui Li, Fangyun Wei, Chao Zhang, and Hongyang Zhang. 2024.
\newblock \href {https://arxiv.org/abs/2406.16858} {Eagle-2: Faster inference of language models with dynamic draft trees}.
\newblock \emph{Preprint}, arXiv:2406.16858.

\bibitem[{Lin et~al.(2025)Lin, Ji, Zhai, Shen, Zhang, Fang, and Gao}]{lin2025lifecycleroutingvulnerabilitiesllm}
Qiqi Lin, Xiaoyang Ji, Shengfang Zhai, Qingni Shen, Zhi Zhang, Yuejian Fang, and Yansong Gao. 2025.
\newblock \href {https://arxiv.org/abs/2503.08704} {Life-cycle routing vulnerabilities of llm router}.
\newblock \emph{Preprint}, arXiv:2503.08704.

\bibitem[{Lin et~al.(2024{\natexlab{a}})Lin, Trivedi, and Sun}]{lin-etal-2024-contextualized}
Zhen Lin, Shubhendu Trivedi, and Jimeng Sun. 2024{\natexlab{a}}.
\newblock \href {https://doi.org/10.18653/v1/2024.emnlp-main.578} {Contextualized sequence likelihood: Enhanced confidence scores for natural language generation}.
\newblock In \emph{Proceedings of the 2024 Conference on Empirical Methods in Natural Language Processing}, pages 10351--10368, Miami, Florida, USA. Association for Computational Linguistics.

\bibitem[{Lin et~al.(2024{\natexlab{b}})Lin, Trivedi, and Sun}]{lin2024generating}
Zhen Lin, Shubhendu Trivedi, and Jimeng Sun. 2024{\natexlab{b}}.
\newblock \href {https://openreview.net/forum?id=DWkJCSxKU5} {Generating with confidence: Uncertainty quantification for black-box large language models}.
\newblock \emph{Transactions on Machine Learning Research}.

\bibitem[{Matarazzo and Torlone(2025)}]{matarazzo2025surveyllm}
Andrea Matarazzo and Riccardo Torlone. 2025.
\newblock \href {https://doi.org/10.48550/arXiv.2501.04040} {A survey on large language models with some insights on their capabilities and limitations}.
\newblock \emph{arXiv preprint arXiv:2501.04040}.
\newblock Version 2.

\bibitem[{Ong et~al.(2025)Ong, Almahairi, Wu, Chiang, Wu, Gonzalez, Kadous, and Stoica}]{ong2025routellm}
Isaac Ong, Amjad Almahairi, Vincent Wu, Wei-Lin Chiang, Tianhao Wu, Joseph~E. Gonzalez, M~Waleed Kadous, and Ion Stoica. 2025.
\newblock \href {https://openreview.net/forum?id=8sSqNntaMr} {Route{LLM}: Learning to route {LLM}s from preference data}.
\newblock In \emph{The Thirteenth International Conference on Learning Representations}.

\bibitem[{Qiu and Miikkulainen(2024)}]{qiu2024semantic}
Xin Qiu and Risto Miikkulainen. 2024.
\newblock \href {https://openreview.net/forum?id=LOH6qzI7T6} {Semantic density: Uncertainty quantification for large language models through confidence measurement in semantic space}.
\newblock In \emph{The Thirty-eighth Annual Conference on Neural Information Processing Systems}.

\bibitem[{Ram{\'\i}rez et~al.(2024)Ram{\'\i}rez, Birch, and Titov}]{rez2024optimising}
Guillem Ram{\'\i}rez, Alexandra Birch, and Ivan Titov. 2024.
\newblock \href {https://openreview.net/forum?id=T9cOYH0wGF} {Optimising calls to large language models with uncertainty-based two-tier selection}.
\newblock In \emph{First Conference on Language Modeling}.

\bibitem[{Shafran et~al.(2025)Shafran, Schuster, Ristenpart, and Shmatikov}]{shafran2025reroutingllmrouters}
Avital Shafran, Roei Schuster, Thomas Ristenpart, and Vitaly Shmatikov. 2025.
\newblock \href {https://arxiv.org/abs/2501.01818} {Rerouting llm routers}.
\newblock \emph{Preprint}, arXiv:2501.01818.

\bibitem[{She et~al.(2025)She, Zheng, Liu, Wang, Xing, Yao, and Ho}]{she2025tokenlevelroutinginference}
Jianshu She, Wenhao Zheng, Zhengzhong Liu, Hongyi Wang, Eric Xing, Huaxiu Yao, and Qirong Ho. 2025.
\newblock \href {https://arxiv.org/abs/2504.07878} {Token level routing inference system for edge devices}.
\newblock \emph{Preprint}, arXiv:2504.07878.

\bibitem[{Sriramanan et~al.(2024)Sriramanan, Bharti, Sadasivan, Saha, Kattakinda, and Feizi}]{sriramanan2024llmcheck}
Gaurang Sriramanan, Siddhant Bharti, Vinu~Sankar Sadasivan, Shoumik Saha, Priyatham Kattakinda, and Soheil Feizi. 2024.
\newblock \href {https://openreview.net/forum?id=LYx4w3CAgy} {{LLM}-check: Investigating detection of hallucinations in large language models}.
\newblock In \emph{The Thirty-eighth Annual Conference on Neural Information Processing Systems}.

\bibitem[{Stripelis et~al.(2024)Stripelis, Hu, Zhang, Xu, Shah, Jin, Yao, Avestimehr, and He}]{stripelis2024tensoroperaroutermultimodelrouter}
Dimitris Stripelis, Zijian Hu, Jipeng Zhang, Zhaozhuo Xu, Alay~Dilipbhai Shah, Han Jin, Yuhang Yao, Salman Avestimehr, and Chaoyang He. 2024.
\newblock \href {https://arxiv.org/abs/2408.12320} {Tensoropera router: A multi-model router for efficient llm inference}.
\newblock \emph{Preprint}, arXiv:2408.12320.

\bibitem[{Su et~al.(2025)Su, Lin, Feng, Zheng, Wang, Xiao, Zhao, Liu, Cheng, and Wang}]{su2025cprouteruncertaintyawarerouterllm}
Jiayuan Su, Fulin Lin, Zhaopeng Feng, Han Zheng, Teng Wang, Zhenyu Xiao, Xinlong Zhao, Zuozhu Liu, Lu~Cheng, and Hongwei Wang. 2025.
\newblock \href {https://arxiv.org/abs/2505.19970} {Cp-router: An uncertainty-aware router between llm and lrm}.
\newblock \emph{Preprint}, arXiv:2505.19970.

\bibitem[{Sun et~al.(2025)Sun, Pickett, Nain, and Jones}]{sun2025transformer}
Qi~Sun, Marc Pickett, Ashish~Kumar Nain, and Luke Jones. 2025.
\newblock Transformer layers as painters.
\newblock \url{https://arxiv.org/abs/2407.09298}.

\bibitem[{Taori et~al.(2023)Taori, Gulrajani, Zhang, Dubois, Li, Guestrin, Liang, and Hashimoto}]{alpaca}
Rohan Taori, Ishaan Gulrajani, Tianyi Zhang, Yann Dubois, Xuechen Li, Carlos Guestrin, Percy Liang, and Tatsunori~B. Hashimoto. 2023.
\newblock Stanford alpaca: An instruction-following llama model.
\newblock \url{https://github.com/tatsu-lab/stanford_alpaca}.

\bibitem[{Wang et~al.(2025)Wang, Tan, Chen, Zhou, Chen, and Li}]{wang2025anymaccascadingflexiblemultiagent}
Song Wang, Zhen Tan, Zihan Chen, Shuang Zhou, Tianlong Chen, and Jundong Li. 2025.
\newblock \href {https://arxiv.org/abs/2506.17784} {Anymac: Cascading flexible multi-agent collaboration via next-agent prediction}.
\newblock \emph{Preprint}, arXiv:2506.17784.

\bibitem[{Wang et~al.(2024)Wang, Ma, Zhang, Ni, Chandra, Guo, Ren, Arulraj, He, Jiang et~al.}]{wang2024mmlupro}
Yubo Wang, Xueguang Ma, Ge~Zhang, Yuansheng Ni, Abhranil Chandra, Shiguang Guo, Weiming Ren, Aaran Arulraj, Xuan He, Ziyan Jiang, and 1 others. 2024.
\newblock \href {https://arxiv.org/abs/2406.01574} {Mmlu-pro: A more robust and challenging multi-task language understanding benchmark}.
\newblock In \emph{Advances in Neural Information Processing Systems (NeurIPS)}.

\bibitem[{Wu et~al.(2024)Wu, Bansal, Zhang, Wu, Li, Zhu, Jiang, Zhang, Zhang, Liu, Awadallah, White, Burger, and Wang}]{wu2024autogen}
Qingyun Wu, Gagan Bansal, Jieyu Zhang, Yiran Wu, Beibin Li, Erkang Zhu, Li~Jiang, Xiaoyun Zhang, Shaokun Zhang, Jiale Liu, Ahmed~Hassan Awadallah, Ryen~W White, Doug Burger, and Chi Wang. 2024.
\newblock \href {https://openreview.net/forum?id=BAakY1hNKS} {Autogen: Enabling next-gen {LLM} applications via multi-agent conversations}.
\newblock In \emph{First Conference on Language Modeling}.

\bibitem[{Xu et~al.(2025)Xu, Jiang, Niu, Deng, Poovendran, Choi, and Lin}]{xu2024magpie}
Zhangchen Xu, Fengqing Jiang, Luyao Niu, Yuntian Deng, Radha Poovendran, Yejin Choi, and Bill~Yuchen Lin. 2025.
\newblock \href {https://arxiv.org/abs/2406.08464} {Magpie: Alignment data synthesis from scratch by prompting aligned llms with nothing}.
\newblock In \emph{International Conference on Learning Representations (ICLR)}.

\bibitem[{Yu et~al.(2025)Yu, Goudarzi, and Toosi}]{yu2025efficientroutinginferencerequests}
Shibo Yu, Mohammad Goudarzi, and Adel~Nadjaran Toosi. 2025.
\newblock \href {https://arxiv.org/abs/2507.15553} {Efficient routing of inference requests across llm instances in cloud-edge computing}.
\newblock \emph{Preprint}, arXiv:2507.15553.

\bibitem[{Zhang et~al.(2025{\natexlab{a}})Zhang, Mehradfar, Dimitriadis, and Avestimehr}]{zhang2025leveraginguncertaintyestimationefficient}
Tuo Zhang, Asal Mehradfar, Dimitrios Dimitriadis, and Salman Avestimehr. 2025{\natexlab{a}}.
\newblock \href {https://arxiv.org/abs/2502.11021} {Leveraging uncertainty estimation for efficient llm routing}.
\newblock \emph{Preprint}, arXiv:2502.11021.

\bibitem[{Zhang et~al.(2025{\natexlab{b}})Zhang, Zhan, and Ye}]{zhang2025capabilityinstructiontuningnew}
Yi-Kai Zhang, De-Chuan Zhan, and Han-Jia Ye. 2025{\natexlab{b}}.
\newblock \href {https://arxiv.org/abs/2502.17282} {Capability instruction tuning: A new paradigm for dynamic llm routing}.
\newblock \emph{Preprint}, arXiv:2502.17282.

\bibitem[{Zhao et~al.(2023)Zhao, Zhou, Li, Tang, Wang, Hou, Min, Zhang, Zhang, Dong, Du, Yang, Chen, Chen, Jiang, Ren, Li, Tang, Liu, Liu, Nie, and Wen}]{zhao2023surveyllm}
Wayne~Xin Zhao, Kun Zhou, Junyi Li, Tianyi Tang, Xiaolei Wang, Yupeng Hou, Yingqian Min, Beichen Zhang, Junjie Zhang, Zican Dong, Yifan Du, Chen Yang, Yushuo Chen, Zhipeng Chen, Jinhao Jiang, Ruiyang Ren, Yifan Li, Xinyu Tang, Zikang Liu, and 3 others. 2023.
\newblock \href {https://doi.org/10.48550/arXiv.2303.18223} {A survey of large language models}.
\newblock \emph{arXiv preprint arXiv:2303.18223}.
\newblock Version 16, last revised March 2025.

\bibitem[{Zhao et~al.(2024)Zhao, Jin, and Mao}]{zhao2024eagleefficienttrainingfreerouter}
Zesen Zhao, Shuowei Jin, and Z.~Morley Mao. 2024.
\newblock \href {https://arxiv.org/abs/2409.15518} {Eagle: Efficient training-free router for multi-llm inference}.
\newblock \emph{Preprint}, arXiv:2409.15518.

\end{thebibliography}

\clearpage
\newpage

\appendix

\section{Implementation Details}
All experiments are conducted with a fixed 
random seed (seed=42) to ensure reproducibility. Due to computational 
constraints, we report single-run results for all experiments.

\section{Benchmark Datasets}
\label{app:datasets}

We utilize six datasets spanning general instruction following, reasoning, and domain-specific knowledge. Table~\ref{tab:dataset_stats} summarizes the statistics of each dataset.

\begin{itemize}
    \item \textbf{In-Domain:} 
    We use \textit{Alpaca}~\citep{alpaca} for general instruction tuning. For knowledge-intensive tasks, we incorporate \textit{MMLU} ~\citep{hendrycks2021measuring}. Mathematical reasoning capabilities are represented by \textit{Big-Math}~\citep{albalak2025bigmath} .
    \item \textbf{Out-of-Domain:} 
    To evaluate generalization, we employ \textit{Magpie}~\citep{xu2024magpie} for aligned dialogue scenarios. For complex knowledge evaluation, we use \textit{MMLU Pro}~\citep{wang2024mmlupro}, which extends MMLU with harder distractors and broader subject coverage. \textit{MATH}~\citep{hendrycks2021math} is used to assess advanced problem-solving skills not covered in the training distribution.
\end{itemize}

\begin{table}[h]
    \centering
    \small
    \caption{Benchmark statistics for router training and evaluation.}
    \label{tab:dataset_stats}
    \begin{tabular}{lccc}
    \toprule
    \textbf{Dataset} & \textbf{Domain} & \textbf{Train/Val} & \textbf{Test} \\
    \midrule
    Alpaca & General & 3.2K/0.8K & 1K \\
    MMLU & Knowledge & 3.2K/0.8K & 10K \\
    Big Math & Math & 3.2K/0.8K & 1K \\ 
    \midrule
    Magpie & General & --- & 10K \\
    MMLU-Pro & Knowledge & --- & 12K \\
    MATH & Math & --- & 5K \\
    \bottomrule
    \end{tabular}
\end{table}

\subsection{Ground Truth Label Construction}
\label{app:label_construction}

\paragraph{Exact Reasoning Tasks.}
For tasks requiring precise reasoning or factual correctness, rule-based string 
matching is often brittle due to format variations. To ensure robust evaluation, 
we leverage \textbf{xVerify},\footnote{\url{https://github.com/IAAR-Shanghai/xVerify}} 
a specialized open-source verification framework, specifically the \texttt{xVerify-9B-C} 
model. Given the query and the small model's response, xVerify performs semantic 
parsing and verification against the ground truth, outputting a hard binary 
correctness label:
\begin{equation}
y = \text{xVerify}(q, r_{\text{small}}, a_{\text{gold}}),
\end{equation}
where $y=1$ indicates correctness (no routing needed) and $y=0$ indicates failure 
(route to large model).

\paragraph{Open-ended Generation Tasks.}
For instruction-following tasks without unique answers, we use GPT-5 as an 
LLM-as-a-Judge evaluator\footnote{As a proxy for SOTA performance. GPT-5 also 
serves as our large model; as judge, it blindly scores all responses without 
knowledge of their source.} to score responses from 0 to 10. For each query $q$, 
we compare the small model's score $S_{\text{small}}$ against the SOTA score 
$S_{\text{sota}}$ (prompt in Figure~\ref{fig:judge_prompt}):
\begin{equation}
y = \mathbb{1}(S_{\text{small}} \ge S_{\text{sota}}).
\end{equation}
This yields $y=1$ (no routing needed) when the small model performs comparably, 
and $y=0$ (route to large model) otherwise.
\begin{figure}[h]
    \centering
    \begin{tcolorbox}[
        colback=blue!5!white,
        colframe=blue!75!black,
        title=Example Query,
        sharp corners,
        enhanced,
        boxrule=0.7pt,
        width=0.98\linewidth,
        arc=2mm,
        left=2pt,
        right=2pt,
        top=2pt,
        bottom=2pt
        ]
        \footnotesize
        \textbf{System Prompt:} You are a helpful assistant.

        \textbf{Instruction:} Please act as an impartial judge and evaluate the quality of the response provided by an AI assistant to the user question displayed below. Your evaluation should consider factors such as the helpfulness, relevance, accuracy, depth, creativity, and level of detail of the response.

        Begin your evaluation by providing a short explanation. Be as objective as possible. After providing your explanation, you must rate the response on a scale of 1 to 10 by strictly following this format: ``[[rating]]'', for example: ``Rating: [[5]]''.

        \vspace{3pt}
        \hrule
        \vspace{3pt}
        \textbf{[Question]}\\
        \{question\}

        \textbf{[The Start of Assistant's Answer]}\\
        \{answer\}\\
        \textbf{[The End of Assistant's Answer]}
    \end{tcolorbox}
    \caption{The prompt template used for LLM-as-a-Judge evaluation on open-ended generation tasks (e.g., AlpacaEval, Magpie). Both the small model and the SOTA proxy model responses are scored using this template to construct the relative ground truth labels.}
    \label{fig:judge_prompt}
\end{figure}

\section{Pseudocode for ProbeDirichlet}

\begin{algorithm}[H]
\caption{ProbeDirichlet}
\label{alg:dynamic_fusion}
\begin{algorithmic}[1]
\Procedure{Forward}{$H \in \mathbb{R}^{B \times L \times D}$, return\_uncertainty}
    \If{probe\_type = "softmax"}
        \State $w = \text{softmax}(\theta_w)$ \Comment{Fixed layer weights}
        \State $F = \sum_{l=1}^{L} H[:, l, :] \cdot w[l]$ 
        \State \Return $\text{Linear}(F)$, None
    \ElsIf{probe\_type = "dirichlet"}
        \State $\alpha = e^{\beta_0} \cdot \text{softmax}(\theta_{\alpha})$ \Comment{Concentration params}
        \If{training}
            \State $w \sim \text{Dirichlet}(\alpha)$ \Comment{Sample weights}
            \State $u = -\sum_{l} w_l \log w_l$ \Comment{Entropy uncertainty}
        \Else
            \State $w = \alpha / \sum_l \alpha_l$ \Comment{Expected weights}
            \State $u = \log(\sum_l \alpha_l)$ \Comment{Total concentration}
        \EndIf
        \State $F = \sum_{l=1}^{L} H[:, l, :] \cdot w[:, l, :]$
        \State \Return $\text{Linear}(F)$, $u$
    \EndIf
\EndProcedure
\end{algorithmic}
\end{algorithm}

\textbf{Mean Pooling:}
$$\hat{z}(x) = \frac{1}{L}\sum_{l=1}^{L} z^{(l)}(x)$$

\section{Supplemental Experiments}
\subsection{Layer Importance Analysis}
\label{app:layer_analysis}

\begin{figure}[h]
    \centering
    \includegraphics[width=0.9\columnwidth]{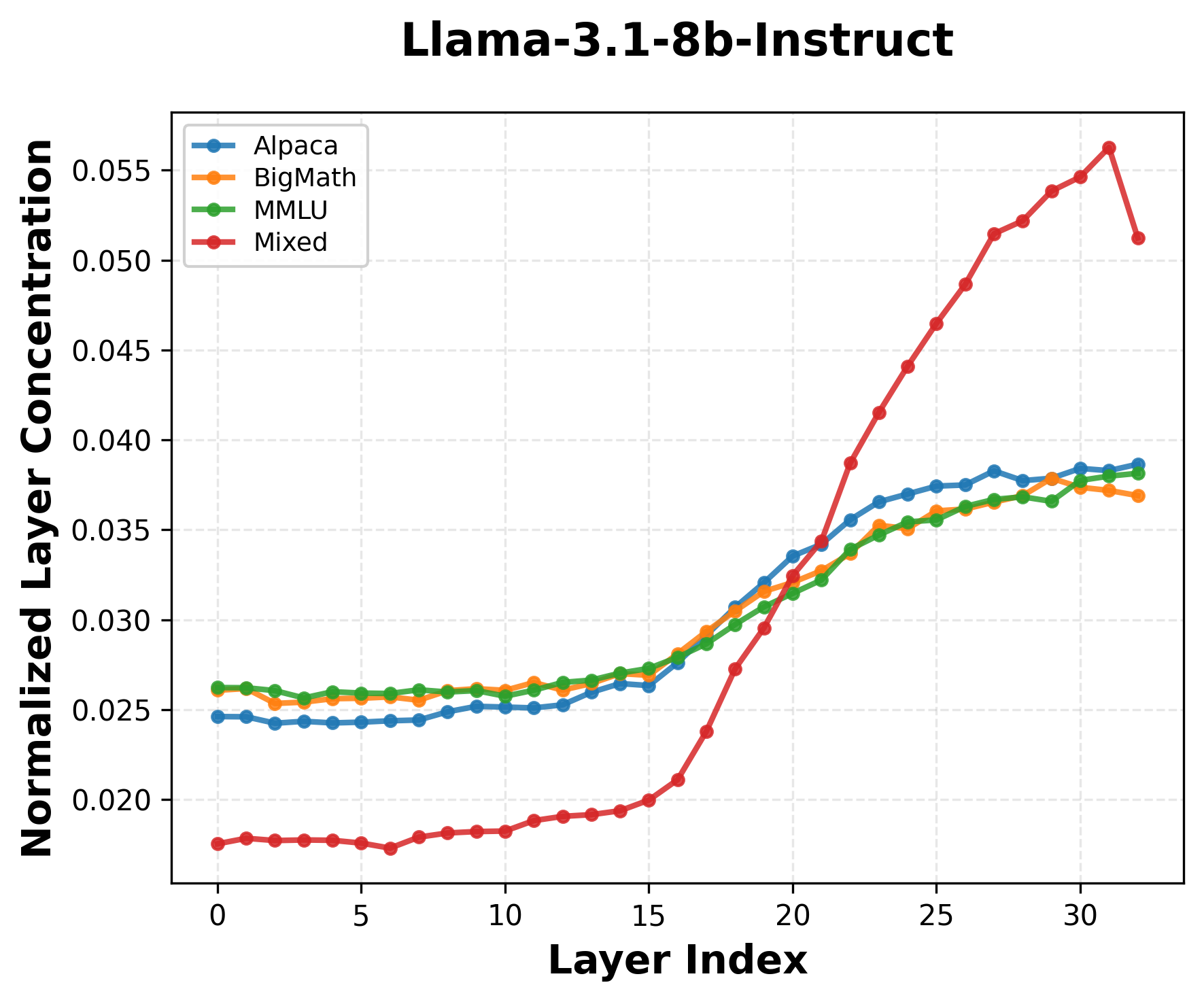}
    \caption{Normalized layer concentration across different training datasets. 
    Deeper layers show higher importance, especially for mixed data.}
    \label{fig:layer_concentration}
\end{figure}

To understand how training data affects layer importance, we visualize the 
normalized layer concentration for Llama-3.1-8b-Instruct in 
Figure~\ref{fig:layer_concentration}. Across all training datasets, deeper 
layers show higher concentration, with the mixed dataset 
exhibiting the most pronounced pattern. Combined with our 
earlier analysis on data diversity, this suggests that deeper layers encode 
stronger signals about the model's capability to answer a given query, making 
them particularly informative for routing decisions.

\subsection{When Routing is Not Enough: A Case Study}
\label{app:case_study}
To illustrate both the effectiveness and limitations of routing systems, 
we analyze queries where our router correctly identified difficulty but the 
strong model still failed. Consider the following example:

\begin{tcolorbox}[colback=blue!5!white, colframe=blue!75!black, title=Example Query]
\textbf{Query:} This biome has cold winters and is known for its pine forests.

\textit{Options:} 
A. Tundra \quad B. Rainforest \quad C. Grassland \quad D. Chaparral \quad E. Savanna\\
F. Alpine \quad G. Wetland \quad H. Deciduous forests \quad I. Desert \quad J. Taiga

\textbf{Small Model:} \boxed{\text{J}} \\
\quad \textbf{Large Model:} \boxed{\text{J}} \\
\quad \textbf{Ground Truth:} \boxed{\text{H}}
\end{tcolorbox}
In such cases, routing becomes ineffective: both models converge on the 
same incorrect answer, making it futile whether the system routes to save 
cost or to seek quality.This reveals critical gaps in current routing frameworks. When both models 
fail on the same query, the system faces a fundamental choice: it can route 
to the small model to save cost, but this delivers incorrect results that may 
mislead users; or route to the large model, which wastes resources without 
improving quality. 

Addressing this requires two complementary strategies. The model pool should include more capable or specialized alternatives to 
handle queries where current models fail. Equally important, routing frameworks must incorporate 
uncertainty-aware mechanisms to detect when no available model is confident in 
these cases, the system should explicitly communicate uncertainty to users, rather than defaulting to the small model to save cost 
while silently delivering incorrect results.

\end{document}